\documentclass{article}

\usepackage[main,final]{neurips_2026}

\usepackage[utf8]{inputenc} %
\usepackage[T1]{fontenc}    %
\usepackage{hyperref}       %
\usepackage{url}            %
\usepackage{booktabs}       %
\usepackage{amsfonts}       %
\usepackage{nicefrac}       %
\usepackage{microtype}      %
\usepackage{xcolor}         %
\usepackage{bbm}
\usepackage{graphicx}
\usepackage{wrapfig}
\usepackage{subcaption}
\usepackage{amsmath}
\usepackage{amssymb}
\usepackage{amsthm}
\usepackage[normalem]{ulem} %
\hypersetup{hidelinks}

\usepackage[most]{tcolorbox}
\tcbuselibrary{listings}
\usepackage{enumitem}
\usepackage{placeins} %
\usepackage{xurl}     %

\definecolor{aggcolor}{HTML}{75D19F}
\definecolor{promptcolor}{HTML}{2C3E50}

\newtcolorbox{callout}[1][]{%
  enhanced,
  colback=aggcolor!10!white,
  colframe=aggcolor,
  fonttitle=\bfseries\large,
  coltitle=aggcolor!60!black,
  title=#1,
  detach title,
  before upper={\tcbtitle\par\smallskip},
  arc=3pt,
  boxrule=0.6pt,
  left=12pt, right=12pt, top=8pt, bottom=8pt,
}

\newtcolorbox{calloutinline}[1][]{%
  enhanced,
  colback=aggcolor!10!white,
  colframe=aggcolor,
  arc=3pt,
  boxrule=0.6pt,
  left=12pt, right=12pt, top=6pt, bottom=6pt,
  before upper={\textcolor{aggcolor!60!black}{\bfseries #1.}~},
}

\newtcblisting{promptbox}[2][]{%
  listing only,
  listing options={
    basicstyle=\ttfamily\footnotesize,
    breaklines=true,
    breakatwhitespace=true,
    columns=fullflexible,
    keepspaces=true,
    showstringspaces=false,
    upquote=true,
    aboveskip=0pt, belowskip=0pt,
  },
  enhanced,
  breakable,
  colback=promptcolor!4!white,
  colframe=promptcolor!70!white,
  fonttitle=\bfseries\sffamily\small,
  coltitle=promptcolor,
  title={#2},
  detach title,
  before upper={\tcbtitle\par\smallskip},
  arc=2pt,
  boxrule=0.4pt,
  left=8pt, right=8pt, top=8pt, bottom=8pt,
  #1,
}

\newtheorem{proposition}{Proposition}

\newtheorem{lemma}{Lemma}

\theoremstyle{definition}
\newtheorem{definition}{Definition}

\title{Vector Policy Optimization:\\ Training for Diversity Improves Test-Time Search}

\author{%
  Ryan Bahlous-Boldi\textsuperscript{1,2} \quad
  Isha Puri\textsuperscript{1} \quad
  Idan Shenfeld\textsuperscript{1,2} \quad
  Akarsh Kumar\textsuperscript{1}  \quad
  Mehul Damani\textsuperscript{1} \\[0.5em]
  \textbf{Sebastian Risi\textsuperscript{4}}  \quad
  \textbf{Omar Khattab\textsuperscript{1}}  \quad
  \textbf{Zhang-Wei Hong\textsuperscript{1,2,3}}  \quad
  \textbf{Pulkit Agrawal\textsuperscript{1,2}} 
  \\[0.9em]
  \textsuperscript{1}MIT \quad
  \textsuperscript{2}Improbable AI Lab \quad 
  \quad \textsuperscript{3}MIT-IBM Computing Research Lab \quad \textsuperscript{4}Sakana AI
}

\begin{document}

\maketitle

\vspace{-1em}

\begin{abstract}
Language models must now generalize out of the box to novel environments and work inside inference-scaling search procedures, such as AlphaEvolve, that select rollouts with a variety of task-specific reward functions. Unfortunately, the standard paradigm of LLM post-training optimizes a pre-specified scalar reward, often leading current LLMs to produce low-entropy response distributions and thus to struggle at displaying the diversity that inference-time search will require. We propose \textbf{Vector Policy Optimization (VPO)}, an RL algorithm that explicitly trains policies to anticipate diverse downstream reward functions and to produce diverse solutions. VPO exploits that rewards are often vector-valued in practice, like per-test-case correctness in code generation or, say, multiple different user personas or reward models.
VPO combines multi-answer prompting with a set-level reward to train the LLM to produce sets of solutions that specialize to different trade-offs in the vector reward space.
Across four tasks, VPO matches or beats the strongest scalar RL baselines on test-time search (e.g.\ pass@$k$ and best@$k$), with the gap widening as the search budget grows.
For evolutionary search, VPO models unlock problems that GRPO models cannot solve at all.
As test-time search becomes standard, diversity may need to become the default post-training objective.

\end{abstract}

\begin{figure}[h]
    \centering
    \includegraphics[width=0.99\linewidth]{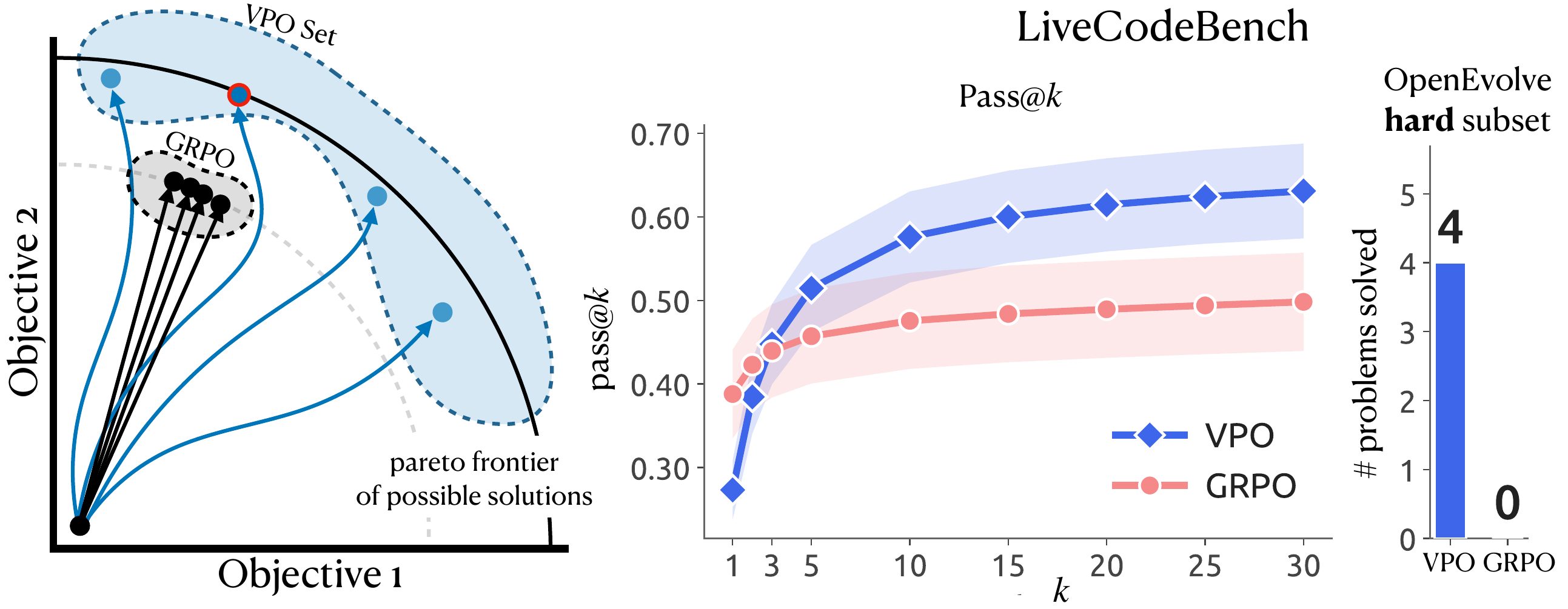}
    \caption{\textbf{Vector Policy Optimization (VPO).} When maximizing a scalar, GRPO sends all solutions to the same, potentially sub-optimal, solution. VPO simultaneously optimizes across different reward weightings, increasing the chance of finding a superior target solution. As a result, on LiveCodeBench, for example, VPO results in better test-time search performance, whether using pass@$k$ or even complex evolutionary test-time search like AlphaEvolve.}
    \label{fig:teaser}
\end{figure}

\section{Introduction}

Many modern language model pipelines are a hybrid of a learned policy with explicit test-time search, ranging from simple rejection sampling with a verifier \citep{cobbe2021training, brown2024large} to complex evolutionary methods like AlphaEvolve \citep{novikov2025alphaevolve,lange2025shinkaevolve} to agentic systems that plan and backtrack across long horizons. In all these settings, search selects candidates from a pool of solutions generated by the learned policy. The quality of this pool is therefore a bottleneck for down-stream search performance. However, existing RL post-training methods actively undermine it: policy gradient methods like GRPO \citep{shao2024deepseekmath} drive the policy toward a narrow set of high-probability responses \citep{wu2025invisible,yue2025does}, so that after training,  additional samples become near-duplicates \citep{gx2025kl,kirk2024understanding}. This mode collapse is precisely what makes downstream search ineffective.  \textbf{We argue that the role of RL post-training should not be to converge on a single best response, but to maximize the diversity of a set of competent solutions, leaving exploitation entirely to search}.

To train a policy that produces diverse yet competent solutions, we exploit the fact that, in many realistic tasks, rewards are a composition of many sub-components: per-test-case correctness for code generation, per-criterion ratings for RLHF, or per-sub-question success in multi-hop reasoning.  We can treat such rewards as a vector. %

This decomposition provides a natural axis for diversity. Scalar notions of diversity such as output entropy are blind to solution quality and do not naturally push training toward qualitatively distinct strategies. A vector decomposition, by contrast, provides semantically meaningful axes: solutions that excel on different subsets of reward components are behaviorally distinct, or solve the problem in a potentially different way.
Rather than collapsing these components into a single scalar and optimizing toward one peak, we train the model to produce solutions that cover the \textbf{trade-offs} between reward components. Randomizing reward weightings alone does not incentivize diversity; instead, we evaluate each weighting against the best candidate in a set, rewarding specialization across different trade-offs. These responses form the supported part of the Pareto front \citep{roijers2013survey}. They are competent overall, but diverse in which objectives each one prioritizes.

We propose \textbf{Vector Policy Optimization (VPO)}\footnote{Code: \url{https://github.com/ryanboldi/vpo}}, a concrete instantiation of this idea. Concretely, VPO combines multi-answer generation~\citep{puri_reaching_2026} with randomly sampled reward weightings, training the model to produce sets of candidates that span different reward trade-offs rather than collapsing onto a single point. Together, these mechanisms maintain a richer candidate distribution so that test-time search can extract increasingly better solutions as the sample budget grows.

We evaluate VPO across four diverse settings that span multi-hop question answering, logic reasoning, navigation, tool use and coding. In our results, VPO matches or beats the strongest scalar baselines on test-time best@$k$ across our four benchmarks, with the gap widening as the candidate budget grows. The advantage of VPO holds as we scale up: on LiveCodeBench, a VPO-trained Qwen2.5-Coder-7B-Instruct improves both pass@$k$ and best@$k$ over a matched-compute GRPO checkpoint, and inside the OpenEvolve search loop unlocks problems that GRPO cannot solve at all within our search budget (Fig.~\ref{fig:teaser}). %

\section{What kind of diversity are we after?}
\label{sec:prelim}

\paragraph{Motivation}
A downstream search procedure only benefits from diversity if the candidates differ along whatever specific axes the search requires. Surface-level variation, semantic diversity, or noisy sampling are not enough. Search needs a pool of candidates that realizes different high-quality trade-offs between the objectives underlying the task. This becomes important once language models are deployed inside search-augmented systems. At inference time, the model is no longer evaluated one response at a time. Instead, the system generates many candidate solutions and selects among them. In this regime, committing the entire policy to a single trade-off is unnecessarily restrictive. We argue that the goal is no longer to produce one response that is optimal under a single fixed objective, but to produce a \emph{set} of responses that spans multiple plausible trade-offs, so downstream search can choose among them.

We call this property \emph{reward diversity}. A reward-diverse candidate pool contains solutions that are each optimal under different weightings of the underlying reward components. Intuitively, the policy remains deliberately non-committal: instead of collapsing onto a single mode, it preserves multiple strategies that perform well under different preferences.

\paragraph{Setting.}
Let $x$ denote a prompt and $y$ a response sampled from policy $\pi_\theta(\cdot \mid x)$.
In many practical tasks, the reward signal decomposes naturally into $d$ components,
$r(x, y) = [r_1(x,y), \dots, r_d(x,y)] \in \mathbb{R}^d$,
where each $r_i$ captures a distinct aspect of response quality. For example, $r_i$ may be
per-test-case correctness in code generation~\citep{chen2021evaluating},
per-criterion preference scores in RLHF~\citep{wang2024interpretable},
per-hop correctness in multi-hop reasoning~\citep{trivedi2022musique},
or per-tool-call structural and content scores in agentic tasks~\citep{qian2025toolrl}.
Any weighting $w \in \Delta^{d-1}$ over the simplex induces a scalar objective $w^\top r(x,y)$. On the other hand,
standard post-training fixes a single weighting $w^*$ and trains the policy to maximize
$\mathbb{E}_{y \sim \pi_\theta(\cdot \mid x)}\bigl[w^{*\top} r(x, y)\bigr]$.

Under the standard single-response framing, this is a sensible objective. If the policy emits only one answer, and evaluation uses a known weighting $w^*$, then directly optimizing $w^*$ is reasonable. The situation changes once inference-time search is introduced. Search benefits precisely from candidates that realize different trade-offs. A policy trained only under $w^*$ has no incentive to preserve such alternatives, and policy-gradient training instead concentrates probability mass onto whichever strategy currently maximizes the scalarized reward. Additional samples then become increasingly redundant.

Crucially, preserving these alternatives is useful even when the deployment objective $w^*$ is known in advance. The reason is that search operates over \emph{sets} of candidates rather than individual responses. A candidate pool that spans multiple reward trade-offs gives the search procedure more opportunities to discover high-performing solutions under $w^*$ itself.

Under scalar training, optimization commits aggressively toward whichever responses currently score highest under $w^*$. Alternative strategies that sacrifice one component in exchange for another are suppressed early, even if they would eventually lead to stronger solutions under the same final objective. As training progresses, the candidate distribution collapses, and additional samples become increasingly redundant. Reward diversity counteracts this collapse by preserving solutions that are optimal under different regions of the reward simplex. Many of these solutions may appear locally suboptimal under $w^*$, yet still contain partial reasoning patterns, decompositions, or strategies that ultimately yield higher-scoring outcomes under $w^*$ itself. In this sense, reward diversity functions as a structured form of exploration: instead of committing prematurely to a single trade-off, the policy maintains a population of competent alternatives long enough for search to exploit them.

This perspective connects naturally to work in multi-objective RL~\citep{roijers2013survey,Hayes_2022} and to lexicase selection in evolutionary computation~\citep{spector_assessment_2012,spector2024particularity,la_cava_probabilistic_2019,ni2024dalex}. Both preserve solutions that are optimal under different subsets or weightings of objectives rather than collapsing all criteria into a single aggregate score. Importantly, however, our goal is different from classical multi-objective optimization. We do not seek a policy conditioned on user-specified preferences, nor do we assume the deployment objective is unknown. Our goal remains performance under a fixed weighting $w^*$. The key difference is that in a search-augmented regime, the best way to optimize $w^*$ may be to train a policy that maintains reward-diverse candidate sets rather than immediately collapsing onto a single optimum.

\section{Method: Vector Policy Optimization}
\label{sec:method}

\begin{figure}
    \centering
    \includegraphics[width=1\linewidth]{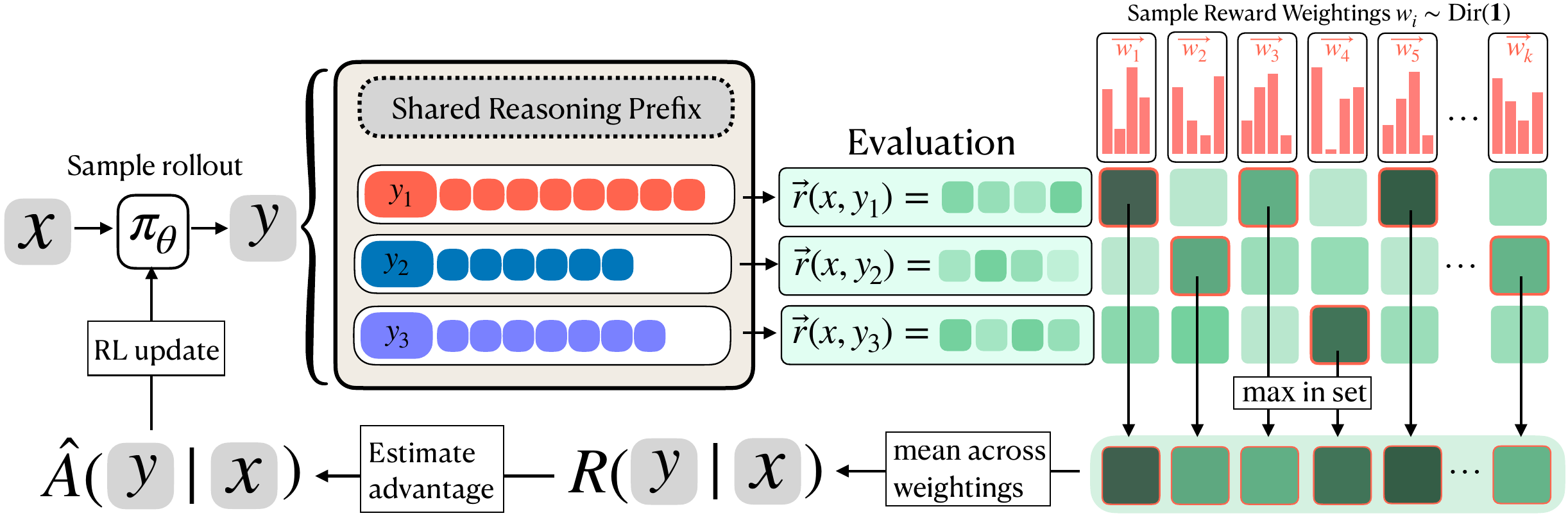}
    \caption{Overview of VPO. Given a prompt $x$, the policy $\pi_\theta$ emits $m$ answers $y_1,\dots,y_m$ in one autoregressive chain, and each answer is scored with a reward vector $r(x,y_i)\in\mathbb{R}^d$. We sample weightings $w\sim\mathrm{Dir}(\mathbf{1})$, take the best answer under each, and average over weightings to obtain the set-level reward $\hat{R}(S)$, which is assigned to the whole chain.}
    \label{fig:method}
\end{figure}

In Section~\ref{sec:prelim}, we argued that the right RL target is reward-diverse sets, or collections of candidates that are each optimal under some weighting of the reward components. This section describes our proposed algorithm, \textbf{Vector Policy Optimization (VPO)}, which trains a policy to produce such sets. VPO has two key components (Fig.~\ref{fig:method}). First, we train a model to generate multiple candidate completions per prompt within a single autoregressive rollout. Second, we replace a fixed reward weighting with a distribution over weights, so the model is incentivized to span its candidates across different trade-offs between the sub-objectives. Together, these components define a set-level objective that rewards the model for producing diverse, high-quality solutions. We describe each component in detail below.

\subsection{Multi-Answer Chains as In-Context Exploration}

Following \citet{puri_reaching_2026}, we train a language model to produce a set of $m$ candidate completions $S=\{y_1, \dots, y_m\}$ within a single rollout. The completions are emitted sequentially, separated by a delimiter token, so when generating $y_i$, the prefix already contains $y_1, \dots, y_{i-1}$.

This fundamentally changes the nature of exploration. Under standard independent sampling, diversity arises only from stochastic decoding applied to a fixed conditional distribution, producing small variations around whichever mode the policy has concentrated on. In multi-answer rollouts, each new candidate can attend to the ones already emitted, giving the model the capacity to recognize which regions of the solution space are covered and steer subsequent candidates towards different ones. Diversity becomes an explicit, in-context mechanism rather than a byproduct of sampling noise.

Importantly, this mechanism provides the \emph{capacity} for diversity, but not a strong \emph{incentive}\footnote{In practice, \citet{puri_reaching_2026} set the reward to zero when the answers in a set are not unique, which adds an incentive against direct duplicates, but the underlying objective still doesn't reward genuine specialization}. Without an appropriate training signal, the model could still collapse to producing near-identical answers. We confirm this empirically in Section~\ref{sec:results}: Multi-RLVR, which combines multi-answer rollouts with a fixed scalar reward, produced sets whose reward-diversity stagnates early in training. The objective defined in the next section supplies this missing incentive by rewarding sets whose elements specialize to different reward trade-offs.

\subsection{Set-Level Optimization via Stochastic Scalarization}
\label{sec:vpo_objective}
To train a policy to output diverse sets, we replace the fixed scalarization prevalent in RL post training with a distribution over scalarizations. For each rollout, we sample weights $w\sim\mathrm{Dir}(\alpha)$, where $\alpha\in\mathbb{R}^d_{>0}$ defines a distribution over the simplex $\Delta^{d-1}$. We use $\alpha=1$ throughout, which results in uniform distribution over the simplex (i.e. the set of $d$-dimensional vectors that sum to 1). We evaluate a set $S=\{y_1, \dots, y_m\}$ under each sampled scalarization by selecting the best performing element:
\begin{equation}
\label{eq:reward}
R(S) = \mathbb{E}_{w\sim\mathrm{Dir}(\alpha)}\left[\max_{y\in S}w^\top r(x, y)\right]
\end{equation}
This objective directly rewards coverage of the reward space: different elements in $S$ are optimal under different samples of $w$. A set that collapses to identical responses performs well under a narrow region of the simplex, while a set that spans multiple trade-offs performs well across many scalarizations. For a fixed $w$, $\max_{y\in S}w^\top r(x,y)$ is the max@$m$ reward of the set \citep{tang2025optimizing,bagirov2025best}, so $R(S)$ averages max@$m$ rewards over random scalarizations.

The two components of VPO are complementary: multi-answer generation alone does not incentivize diversity, while varying scalarizations for a single output can create instability. Together, they form a stable set-level objective that directly rewards diversity.

\paragraph{What does $R$ reward?}

Given a prompt $x$, we can treat all responses $y_i$ as identical if they have the same reward vectors since $R$ cannot distinguish between them. Every possible weighting $w$ over the simplex has a best response, which is the one with a reward vector $r$ that maximizes $w^\top r$. As $w$ moves across the simplex, the best answer changes, so the simplex splits into regions, each ``won" by a single response. We call any response $y\in\mathcal{Y}$ \emph{supported} if it wins such a region, i.e.\ if it is the best choice for some range of $w$. Supported responses are Pareto-optimal, but not every Pareto-optimal response is supported. Let $n_s$ be the number of supported responses (formal definition in App.~\ref{app:theory}).

\begin{proposition}[Optimal sets are diverse]\label{prop:optimal_sets}
Let $w \sim \mathrm{Dir}(\alpha)$ for any $\alpha \in \mathbb{R}^d_{>0}$ and let $S^*$ maximize $R(S)$ over sets of at most $m$ responses. (i) if $m\geq n_s$, $S^*$ contains every supported response, and hence, for every weighting $w$ (including $w^*$), a response that is best under $w$. (ii) if $m < n_s$, $S^*$ consists of $m$ distinct responses, the optimal value of $R$ strictly increases with $m$, and some optimal set consists of supported responses only.
\end{proposition}

A scalar objective is maximized by a single response. Under $R$, adding a supported response that is missing from a set always increases the set's value, and responses that are not supported are never needed. For $m=1$, $R$ reduces to scalar RL under the mean weighting, and for $d=1$ to best-of-$m$ training (proofs and remarks in App.~\ref{app:theory}).

\paragraph{Reward estimation}
As it only specifies a reward calculation, VPO can be combined with any policy-gradient method. In our experiments, we use GRPO  \citep{shao2024deepseekmath}. For each prompt $x$, we sample $G$ rollouts, each producing a set $S^{(g)} = \{y_1^{(g)}, \dots, y_m^{(g)}\}$ of $m$ completions, along with $K$ scalarization weights $w^{(1)}, \dots, w^{(K)} \stackrel{\mathrm{iid}}{\sim} \mathrm{Dir}(\mathbf{1})$ used in common across the group. The per-rollout Monte-Carlo reward is
\[
\hat{R}(S^{(g)}) \;=\; \frac{1}{K} \sum_{k=1}^K \max_{s \in S^{(g)}} w^{(k)\top} r(x, s),
\]
which estimates $R(S)$ from the previous section. The GRPO advantage is then calculated and applied uniformly to every token in the rollout $g$. The $K$ scalarization weights are shared across the $G$ rollouts in the group, so all $G$ sets are evaluated under the same draws of $w$ and are therefore comparable.

\section{Experimental Setup}

\subsection{Evaluation Tasks}
\label{sec:envs}

We evaluate on four domains chosen to span distinct shapes of multi-objective structure:  (i) binary vs.\ continuous reward components, and (ii) hand-crafted vs.\ metric-based reward shapes. In all our experiments, we used $m{=}3$ candidates per multi-answer chain across all domains. Full reward and prompt details are in App.~\ref{app:env_details}.

\paragraph{Maze.}\label{sec:env_maze} A synthetic navigation task in which the model emits, in text, a sequence of moves from a start corner $S$ to a goal corner $E$, collecting gold and diamond items and avoiding lava along the way. We construct the mazes so that item types are forced to trade off against each other and against reaching the exit: the geometry guarantees that no single route can satisfy every reward component. The reward $r \in \mathbb{R}^4$ has one binary completion component and three item/safety terms in $[0,1]$ (gold, diamond, lava-avoidance); the GRPO scalar is the uniform mean. The maze is our controlled testbed where the trade-off is engineered, which lets us isolate the counter-intuitive question of whether training VPO and evaluating under the \emph{same} uniform mean GRPO was directly optimized for still beats GRPO. We train Qwen3-4B and evaluate on 100 held-out mazes.

\paragraph{MuSiQue.}\label{sec:env_musique} A 2--4 hop reading-comprehension benchmark \citep{trivedi2022musique} in which the model selects supporting paragraphs from 20 candidates and emits a final answer. The reward $r \in \mathbb{R}^5$ has four binary citation indicators (one per gold hop) plus a continuous answer-F1 term. The GRPO scalarization weights the final answer $3\times$ more than each hop's correctness to reflect its priority. We train Qwen3-1.7B and evaluate on a 300-question hop-stratified split.

\paragraph{EUREQA.} The 5-hop subset of a chain-reasoning benchmark \citep{li2024deceptive}, where the model back-chains through five relations to identify five masked entities. The reward $r \in \{0,1\}^5$ is binary per-entity and the GRPO scalar is a uniform mean. Unlike MuSiQue's loosely coupled hops, the chain is causally ordered, so the per-hop training signal is informative about which step failed. We train Qwen3-8B and evaluate on a held-out hard split. Due to the limited size of the eval dataset, we averaged over 4 evaluation seeds.

\paragraph{ToolRL.} A function-calling benchmark \citep{qian2025toolrl} of 3{,}920 train and 80 test prompts. The reward $r \in \mathbb{R}^4$ has one binary structural-format component and three continuous F1 dimensions (tool-name, arg-key, arg-value), graded from trivially solvable to schema-precise; the GRPO scalar is a uniform mean. We train Qwen3-1.7B and evaluate on the 80-prompt test split. Due to the limited size of the eval dataset, we averaged over 4 evaluation seeds.

 We build on veRL~\citep{sheng2024hybridflow} with standard outcome-reward GRPO~\citep{shao2024deepseekmath}. The full training details are documented in App.~\ref{app:training_setup}.

\subsection{Baselines}
\label{sec:baselines}

Our baselines are designed to isolate which ingredient of VPO is responsible for improved test-time search. VPO combines two mechanisms: multi-answer generation within a single autoregressive rollout, and stochastic scalarization over a vector-valued reward. We compare against methods that test whether either ingredient, or existing search-aware RL objectives, are sufficient on their own.

\begin{itemize}[leftmargin=*]

\item \textbf{GRPO} \citep{shao2024deepseekmath} (single-answer, scalar reward).
This is the standard RL post-training baseline. Rewards are collapsed into a fixed scalar objective, and the model emits one completion per prompt. This tests whether ordinary scalar RL already produces candidate pools that are useful for test-time search.

\item \textbf{Multi-RLVR} \citep{puri_reaching_2026} (multi-answer, scalar reward).
This baseline trains the model to emit multiple answers in one rollout, but evaluates the set using the same fixed scalar reward. It tests whether multi-answer generation alone is sufficient, or whether a diversity-preserving reward signal is also needed.

\item \textbf{Random-Weighting GRPO} (single-answer, stochastic scalarization).
Here the model still emits one answer per rollout, but the scalarization weights are resampled during training, $w\sim\mathrm{Dir}(\alpha)$. This tests whether randomizing the reward objective alone is enough without set-level optimization.

\item \textbf{Max-at-$k$ Training} \citep{bagirov2025best}.
This baseline directly optimizes an inference-aware best@$k$/max@$k$ objective. It tests whether explicitly training for best-of-$k$ performance is sufficient without requiring reward-diverse candidate sets.

\item \textbf{MaxRL}. 
MaxRL \citep{tajwar2026maximum} is another search-aware RL objective that uses additional sampling compute during training to better approximate maximum-likelihood-style learning from successful rollouts. It tests whether stronger scalar search-aware training objectives can recover the gains of VPO.

\item \textbf{Goal-Conditioned GRPO}.
This baseline conditions the policy on a target scalarization $w$ and trains it to maximize $w^\top r(x,y)$. It tests the natural multi-objective RL alternative: whether diversity is better obtained by asking for different trade-offs explicitly, rather than by producing a reward-diverse set within one rollout.

\end{itemize}

\subsection{Evaluation Metrics}
\label{sec:metrics}

\paragraph{Best@$k$.} Our central metric is the maximum scalarized reward over a pool of $k$ candidates,
\[
    \mathrm{best}@k(x) \;=\; \max_{s \in S_k(x)}\, w^{\star\top}\, r(x, s),
\]
where $w^\star$ is the per-domain GRPO training scalar (Table~\ref{tab:per_domain_scalars}) and $S_k(x)$ is a uniformly random $k$-subset of the $30$ candidates sampled for prompt $x$. 
Multi-answer methods sample $10$ chains of $m=3$, single-answer methods $30$ i.i.d.\ completions, and $S_k$ ignores chain boundaries. The scaling figures pool whole chains (App.~\ref{app:metrics}).

\paragraph{Reward-space diversity.} Alongside best@$k$, we report the average pairwise $L_1$ distance between the \emph{reward vectors} of the completions in the pool, which measures spread in the reward space $\mathbb{R}^d$: high diversity means the candidates realize different reward trade-offs. A pool that collapses to a single mode has very low diversity even when the surface text varies.

\section{Results}
\label{sec:results}

\begin{figure}[t]
    \centering
    \includegraphics[width=\linewidth]{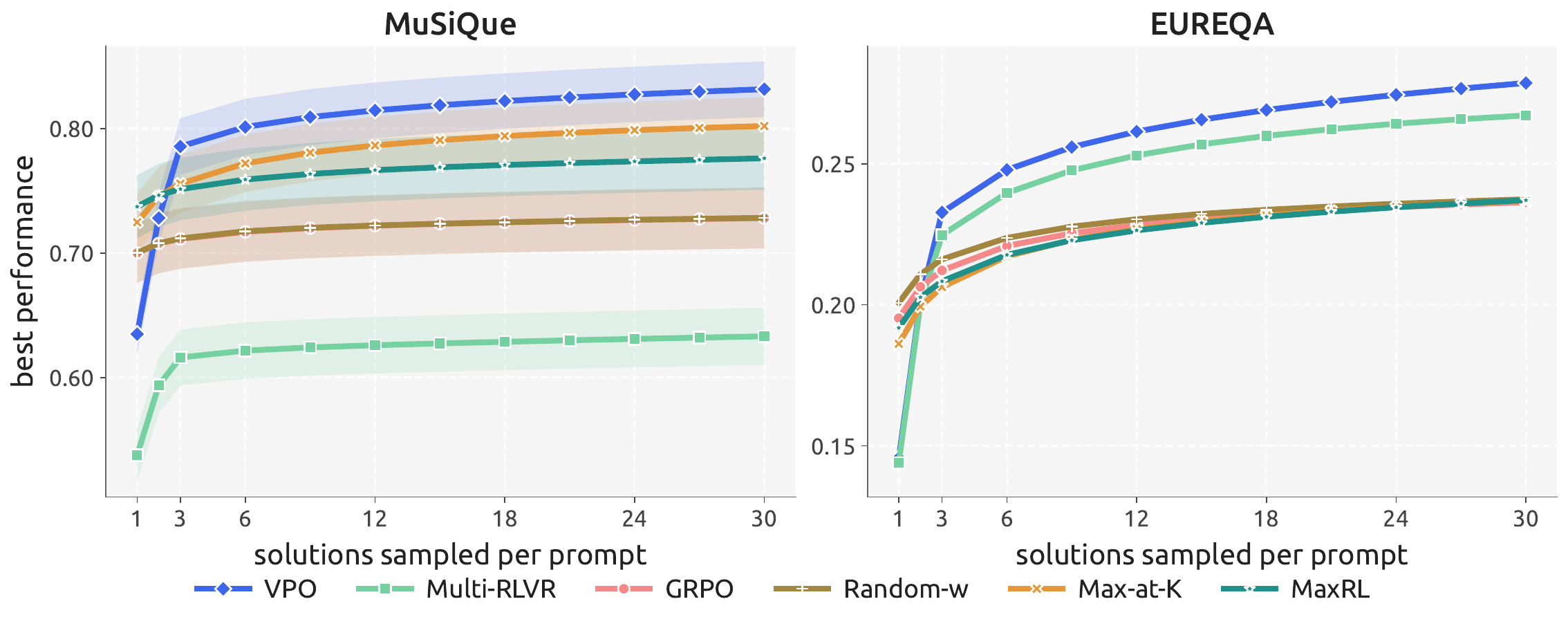}
    \caption{\textbf{Test-time scaling on MuSiQue and EUREQA.} Best@$k$ on the GRPO training scalar as a function of $k$. Scalar GRPO plateaus quickly, reflecting collapse of the candidate pool to near-duplicates, while VPO continues to extract value from additional samples. }
    \label{fig:chain_scaling}
\end{figure}

\paragraph{VPO Improves Gains from Test-Time Search} A central goal of VPO is to improve the effectiveness of test-time search by producing more diverse and useful candidate solutions. We evaluate this through best@$k$, a simple search procedure that chooses the maximum scalarized reward over a pool of $k$ candidates. Across all four domains, Maze (Table~\ref{tab:maze}), MuSiQue (Table~\ref{tab:musique}), EUREQA (Table~\ref{tab:eureqa}), and ToolRL (Table~\ref{tab:toolrl}), VPO improves test-time scaling relative to scalar baselines. While scalar GRPO quickly saturates as $k$ increases, indicating collapsing candidate diversity, VPO continues improving with additional samples and plateaus at a substantially higher level. 
Figure~\ref{fig:chain_scaling} shows this effect on MuSiQue and EUREQA (the companion plot for Maze and ToolRL is in App.~\ref{app:additional_figures}). %

\begin{table}[h]
\centering
\small
\caption{\textbf{Best@$k$ on MuSiQue} (held-out 300-question hop-stratified split). Bold and underline mark the best and second-best trained method in each column (base-model rows excluded). VPO continues improving as $k$ grows while scalar baselines plateau.}
\label{tab:musique}
\begin{tabular}{l c c c c c c}
\toprule
Method & best@$3$ & best@$5$ & best@$10$ & best@$30$ & F1@$30$ & diversity \\
\midrule
Qwen3-1.7B (single-answer prompt) & $0.310$ & $0.339$ & $0.373$ & $0.423$ & $0.161$ & $0.634$ \\
\hspace{1em}$\hookrightarrow$ GRPO & $0.711$ & $0.716$ & $0.721$ & $0.728$ & $0.447$ & $0.054$ \\
\hspace{1em}$\hookrightarrow$ Random-$w$ & $0.712$ & $0.716$ & $0.721$ & $0.728$ & $0.445$ & $0.061$ \\
\hspace{1em}$\hookrightarrow$ Max@$k$ & $\mathbf{0.757}$ & $\underline{0.768}$ & $\underline{0.783}$ & $\underline{0.802}$ & $\underline{0.573}$ & $0.175$ \\
\hspace{1em}$\hookrightarrow$ MaxRL & $\underline{0.751}$ & $0.757$ & $0.765$ & $0.776$ & $0.552$ & $0.068$ \\
\midrule
Qwen3-1.7B (multi-answer prompt) & $0.402$ & $0.439$ & $0.475$ & $0.506$ & $0.379$ & $0.855$ \\
\hspace{1em}$\hookrightarrow$ Multi-RLVR & $0.599$ & $0.616$ & $0.627$ & $0.633$ & $0.498$ & $\mathbf{0.814}$ \\
\hspace{1em}$\hookrightarrow$ VPO (ours) & $0.742$ & $\mathbf{0.780}$ & $\mathbf{0.809}$ & $\mathbf{0.832}$ & $\mathbf{0.678}$ & $\underline{0.587}$ \\
\bottomrule
\end{tabular}

\end{table}

\begin{table}[h]
\centering
\small
\caption{\textbf{Best@$k$ on Maze} (100 held-out mazes). VPO outperforms scalar baselines under the same uniform-mean GRPO scalar even though it was trained on Dirichlet-sampled scalarizations. Bold and underline mark the best and second-best trained method in each column (base-model rows excluded).}
\label{tab:maze}
\begin{tabular}{l c c c c c}
\toprule
Method & best@$3$ & best@$5$ & best@$10$ & best@$30$ & diversity \\
\midrule
Qwen3-4B (single-answer prompt) & $0.341$ & $0.457$ & $0.596$ & $0.714$ & $0.905$ \\
\hspace{1em}$\hookrightarrow$ GRPO & $0.432$ & $0.432$ & $0.432$ & $0.432$ & $0.003$ \\
\hspace{1em}$\hookrightarrow$ Random-$w$ & $0.414$ & $0.420$ & $0.425$ & $0.432$ & $0.090$ \\
\hspace{1em}$\hookrightarrow$ Max@$k$ & $\mathbf{0.526}$ & $\underline{0.552}$ & $\underline{0.568}$ & $\underline{0.577}$ & $\underline{0.671}$ \\
\hspace{1em}$\hookrightarrow$ MaxRL & $0.414$ & $0.426$ & $0.441$ & $0.464$ & $0.206$ \\
\midrule
Qwen3-4B (multi-answer prompt) & $0.066$ & $0.103$ & $0.178$ & $0.334$ & $0.176$ \\
\hspace{1em}$\hookrightarrow$ Multi-RLVR & $0.420$ & $0.430$ & $0.435$ & $0.436$ & $0.187$ \\
\hspace{1em}$\hookrightarrow$ VPO (ours) & $\underline{0.512}$ & $\mathbf{0.564}$ & $\mathbf{0.591}$ & $\mathbf{0.593}$ & $\mathbf{1.006}$ \\
\bottomrule
\end{tabular}

\end{table}

\begin{table}[h]
\centering
\small
\caption{\textbf{Best@$k$ on EUREQA} (held-out half of the hard split, averaged over 4 evaluation seeds). Bold and underline mark the best and second-best trained method in each column (base-model rows excluded).}
\label{tab:eureqa}
\begin{tabular}{l c c c c c}
\toprule
Method & best@$3$ & best@$5$ & best@$10$ & best@$30$ & diversity \\
\midrule
Qwen3-8B (single-answer prompt) & $0.081$ & $0.096$ & $0.117$ & $0.153$ & $0.171$ \\
\hspace{1em}$\hookrightarrow$ GRPO & $0.212$ & $0.219$ & $0.226$ & $0.236$ & $0.119$ \\
\hspace{1em}$\hookrightarrow$ Random-$w$ & $\mathbf{0.216}$ & $0.222$ & $0.229$ & $0.237$ & $0.105$ \\
\hspace{1em}$\hookrightarrow$ Max@$k$ & $0.206$ & $0.214$ & $0.224$ & $0.237$ & $0.140$ \\
\hspace{1em}$\hookrightarrow$ MaxRL & $0.209$ & $0.216$ & $0.224$ & $0.237$ & $0.117$ \\
\midrule
Qwen3-8B (multi-answer prompt) & $0.102$ & $0.113$ & $0.126$ & $0.140$ & $0.171$ \\
\hspace{1em}$\hookrightarrow$ Multi-RLVR & $0.210$ & $\underline{0.230}$ & $\underline{0.249}$ & $\underline{0.267}$ & $\mathbf{0.526}$ \\
\hspace{1em}$\hookrightarrow$ VPO (ours) & $\underline{0.213}$ & $\mathbf{0.236}$ & $\mathbf{0.257}$ & $\mathbf{0.279}$ & $\underline{0.512}$ \\
\bottomrule
\end{tabular}

\end{table}

\begin{table}[h]
\centering
\small
\caption{\textbf{Best@$k$ on ToolRL} (80-prompt held-out split, averaged over 4 evaluation seeds). Bold and underline mark the best and second-best trained method in each column (base-model rows excluded).}
\label{tab:toolrl}
\begin{tabular}{l c c c c c}
\toprule
Method & best@$3$ & best@$5$ & best@$10$ & best@$30$ & diversity \\
\midrule
Qwen3-1.7B (single-answer prompt) & $0.112$ & $0.114$ & $0.116$ & $0.123$ & $0.026$ \\
\hspace{1em}$\hookrightarrow$ GRPO & $\underline{0.921}$ & $0.923$ & $0.924$ & $0.925$ & $0.044$ \\
\hspace{1em}$\hookrightarrow$ Random-$w$ & $0.902$ & $0.904$ & $0.906$ & $0.907$ & $0.064$ \\
\hspace{1em}$\hookrightarrow$ Max@$k$ & $\mathbf{0.940}$ & $\mathbf{0.945}$ & $\underline{0.949}$ & $\mathbf{0.954}$ & $0.131$ \\
\hspace{1em}$\hookrightarrow$ MaxRL & $0.779$ & $0.805$ & $0.826$ & $0.846$ & $\underline{0.510}$ \\
\midrule
Qwen3-1.7B (multi-answer prompt) & $0.093$ & $0.104$ & $0.111$ & $0.116$ & $0.198$ \\
\hspace{1em}$\hookrightarrow$ Multi-RLVR & $0.861$ & $0.883$ & $0.898$ & $0.905$ & $0.308$ \\
\hspace{1em}$\hookrightarrow$ VPO (ours) & $0.897$ & $\underline{0.934}$ & $\mathbf{0.950}$ & $\underline{0.952}$ & $\mathbf{1.297}$ \\
\bottomrule
\end{tabular}

\end{table}

\paragraph{Is the gain due to multi-answer prompting alone?}

We want to understand if generating multiple answers within a shared autoregressive context, as explored by \citet{puri_reaching_2026}, is sufficient for diversity. It appears not. VPO outperformed Multi-RLVR on best@$k$ across all four domains, and the gap widened with $k$, mirroring the pattern against scalar GRPO. The mechanism is visible in App.~\ref{app:additional_figures}, Fig.~\ref{fig:diversity_over_time}, which plots the pairwise $L_1$ distance between the per-rollout \emph{reward vectors} in each candidate pool (which is a measure of how much the pool spreads in reward space, not in token space). Throughout training, Multi-RLVR's reward-space diversity stagnates or remains below VPO on Maze, MuSiQue, and ToolRL. Interestingly, the domain on which Multi-RLVR performs best relative to the other baselines (EUREQA) is also the only one on which its training-time reward-space diversity tracks VPO's, supporting the hypothesis that VPO's training-time diversity is instrumental in improving test-time performance. We conclude that multi-answer prompting supplies the capacity to produce distinct candidates, but under a fixed scalarization, the gradient still pushes every position in the chain toward the same scalar optimum, so the candidates collapse to similar reward vectors. The stochastic scalarization in VPO gives different positions an incentive to specialize.

\paragraph{Is the gain due to more evaluator signal during training or normalization issues?}
\begin{wraptable}{r}{0.45\textwidth}
\vspace{-1.2em}
\centering
\small
\setlength{\tabcolsep}{4pt}
\caption{MuSiQue. Even when GRPO/GDPO receive 3$\times$ the rollouts ($n{=}24$, roughly 3$\times$ the compute), they don't match VPO at $n{=}8$. All results are separate from earlier runs.}
\label{tab:musique-3x-ablation}
\begin{tabular}{l c c}
\toprule
Method & best@3 & $\mathbb{E}_w$[best@3] \\
\midrule
GRPO ($n{=}8$)                       & 0.741 & 0.830 \\
GDPO ($n{=}8$)                       & 0.737 & 0.832 \\
GRPO ($n{=}24$, 3$\times$)           & 0.763 & 0.841 \\
GDPO ($n{=}24$, 3$\times$)           & 0.765 & 0.850 \\
GRPO + rand-$w$ ($n{=}24$)           & 0.755 & 0.839 \\
\textbf{VPO (ours, $n{=}8$)}         & \textbf{0.779} & \textbf{0.856} \\
\bottomrule
\end{tabular}
\vspace{-1em}
\end{wraptable}
Two natural questions arise. First, VPO may benefit simply from receiving 3x more evaluator signal per rollout. %
Second, GRPO is known to be sensitive to reward components with very different variances, where a single high-variance dimension can dominate the advantage \citep{ichihara2025mogrpo}. GDPO \citep{liu_gdpo_2026} addresses this by normalizing the advantage per reward dimension before aggregating. If VPO's gains came from better gradient conditioning across components rather than from set-level diversity, per-dimension normalization should close the gap.

We test both at once on the MuSiQue domain by giving GRPO and GDPO 3x the rollouts ($n=24$) to equalize evaluator calls with VPO at $n=8$. Note that this also gives the baselines 3x the LM rollouts, since their reasoning chains are independent per answer rather than shared, so the comparison is conservative against VPO. Table~\ref{tab:musique-3x-ablation} shows that neither effect explains the gap. At matched $n=8$, GDPO closely tracks GRPO, indicating the per-component normalization is not a binding constraint. At 3x the rollouts, both GRPO and GDPO improve modestly but remain below VPO at $n=8$, and adding random $w$ scalarization at $n=24$ also fails to close the gap. Extra evaluator signal during training and better normalization help marginally, but scalarized objectives still remove the incentive to maintain a diverse candidate set.

\begin{table}[t]
\small
\setlength{\tabcolsep}{4pt}
\begin{minipage}[c]{0.37\textwidth}
\caption{\textbf{Goal-conditioned GRPO on Maze} under fixed $w^*$ and random $w \sim \mathrm{Dir}(\mathbf{1})$. Neither matches VPO on either the gold scalar or the Dirichlet-averaged metric. Bold marks the best entry in each column.}
\label{tab:maze-goalcond}
\end{minipage}\hfill
\begin{minipage}[c]{0.6\textwidth}
\centering
\begin{tabular}{l c c c c}
\toprule
Method & best@$3$ & best@$6$ & $\mathbb{E}_w[\mathrm{best}@3]$ & $\mathbb{E}_w[\mathrm{best}@6]$ \\
\midrule
\multicolumn{5}{l}{Goal-cond.\ GRPO, conditioned on} \\
\hspace{1em}$w = w^\star$ & $0.205$ & $0.205$ & $0.201$ & $0.201$ \\
\hspace{1em}$w \sim \mathrm{Dir}(\mathbf{1})$ & $0.205$ & $0.205$ & $0.201$ & $0.201$ \\
\midrule
VPO (ours) & $\mathbf{0.512}$ & $\mathbf{0.576}$ & $\mathbf{0.512}$ & $\mathbf{0.584}$ \\
\bottomrule
\end{tabular}

\end{minipage}
\end{table}

\paragraph{Why not just condition on $w$?}

To make the model output responses that cover reward space, VPO uses in-context exploration. Prior work instead trains goal-conditioned policies \citep{yang2019generalized,wang2024conditional} that take a target scalarization $w$ as input, optimize $w^\top r$ directly, and create diversity by varying $w$ at inference time. We train a goal-conditioned GRPO policy and condition it either on the canonical $w^*$ or on random $w\sim\mathrm{Dir}(\mathbf{1})$ (Table~\ref{tab:maze-goalcond}). Neither setting matches VPO on Maze, under the gold scalar or under the expected value over random weightings. The goal-conditioned policy mode-collapsed (hence identical columns) and began to ignore the conditioning (hence identical rows), so explicit access to $w$ does not necessarily yield distinct behavior.

\begin{figure}[h]
    \centering
    \includegraphics[width=\linewidth]{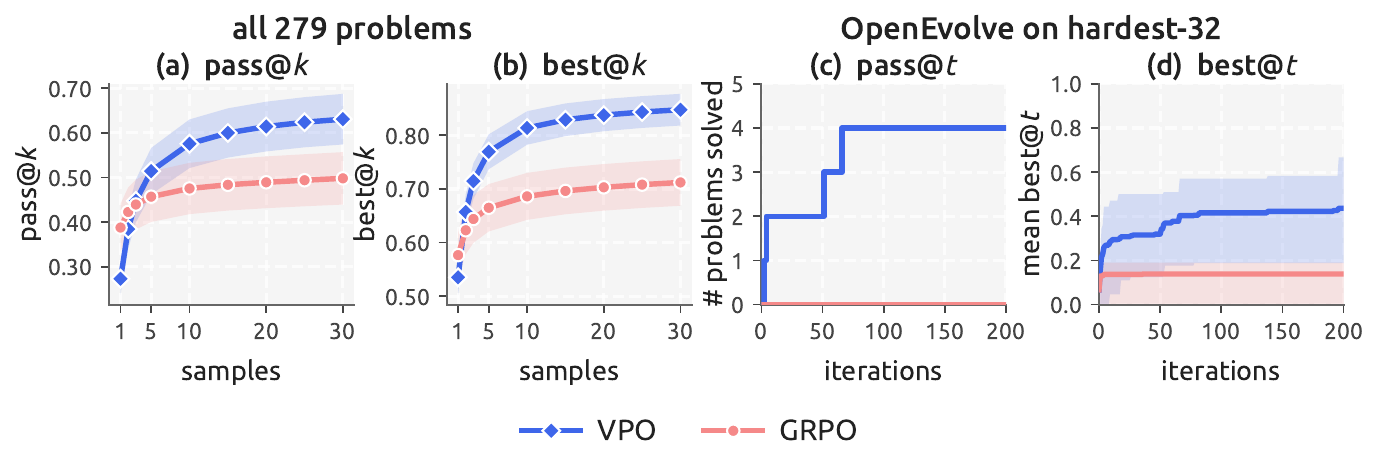}
    \caption{\textbf{LiveCodeBench case study: VPO vs.\ scalar GRPO.} (A) Pass@$k$ on the full $279$-problem held-out split. GRPO is better at $k{=}1$, and VPO overtakes it as $k$ grows. (B) Best@$k$ on the same split. VPO is above GRPO at every $k$. (C, D) Pass@$k$ and best@$k$ over $200$ OpenEvolve iterations on $32$ of the hardest held-out problems, which neither method solved with $30$ candidates each. VPO keeps discovering new solutions while GRPO plateaus.}
    \label{fig:lcb_main_4panel}
\end{figure}

\paragraph{How does VPO scale to harder problems and more sophisticated search?}
\label{sec:livecodebench_case_study}
We compare VPO to scalar GRPO on LiveCodeBench (LCB) \citep{jain2024livecodebench}, a competitive-programming benchmark with a strict temporal held-out split. Both runs train Qwen2.5-Coder-7B-Instruct on the same DeepCoder corpus for one epoch and differ only in prompting, answer parsing, and advantage estimator. GRPO is better at pass@$1$ (Fig.~\ref{fig:lcb_main_4panel}A), where there is no search to benefit from. Under best@$k$ with chains of $m{=}3$, VPO is above GRPO at every $k$, and the gap widens with $k$ (Fig.~\ref{fig:lcb_main_4panel}B). We then run a more capable search, OpenEvolve \citep{openevolve}, on $32$ of the hardest held-out problems, chosen because neither method solved them with $30$ candidates each (App.~\ref{app:env_livecodebench}). Over $200$ iterations VPO keeps discovering new solutions, including problems that neither method solves under best@$k$, while GRPO plateaus early (Fig.~\ref{fig:lcb_main_4panel}C,D). The benefit of diversity grows as the problems get harder and the search gets more capable.

\paragraph{When would VPO not help?}\label{sec:analysis} VPO helps most when the reward components disagree about which response is best. On Maze, MuSiQue, and EUREQA, the components are not collinear under the trained model (App.~\ref{app:ultrafeedback}), many distinct trade-offs are supported, and VPO produces sets that span them. On UltraFeedback \citep{cui2023ultrafeedback} with the five objectives of the ArmoRM reward model \citep{wang2024interpretable}, the components are nearly collinear, so the reward is effectively scalar. There VPO ends at or below scalar baselines on absolute best@$k$, although it still gains the most of any method from best@$1$ to best@$50$.

\section{Related Work}
\label{sec:related_work}

Inference-time search methods such as best-of-$N$ sampling \citep{beirami2025theoretical}, self-consistency \citep{wang2022self}, particle filtering \citep{puri2025probabilistic}, Tree-of-Thoughts \citep{yao2023tree}, and evolutionary search \citep{romera2024mathematical,novikov2025alphaevolve} select among LLM-generated candidates. VPO is a complementary post-training algorithm that trains the policy to produce the diverse, competent pool they rely on. Best-of-$N$ distillation \citep{sessa2025bond,gui2024bonbon}, pass@$k$ and max@$k$ objectives \citep{chow_inference-aware_2025,tang2025optimizing,walder2025pass,chen2025pass,bagirov2025best}, and MaxRL \citep{tajwar2026maximum} also train policies for inference-time selection. These methods optimize performance under a fixed scalar reward, whereas VPO optimizes the expected best reward across a distribution of scalarizations. Additionally, VPO generates candidates sequentially within a single rollout, rather than sampling them independently.

Multi-RLVR \citep{puri_reaching_2026} generates $m$ answers in one rollout under a fixed scalar reward. Polychromic objectives \citep{hamid_polychromic_2025,orney2026poly}, DQO \citep{chen2026dqo}, SetPO \citep{li2026setpo}, and GCPO \citep{chen2026gcpo} encourage diversity among i.i.d.\ samples using external diversity measures. Concurrently, ROSA \citep{gxchen2026usingrewarduncertaintyinduce}, maximizes expected best reward under reward-function uncertainty. ROSA uses i.i.d.\ candidates and per-candidate credit; VPO conditions each candidate on preceding candidates and assigns one advantage to the whole chain. ReMax \citep{koyamada2022remax,nishimori2026emergence} studies how expected-max objectives under reward uncertainty promote stochastic exploration in RL. In text-to-image generation, concurrent multi-axis max@$K$ \citep{onoda2026multiaxis} sums per-objective maxima over i.i.d.\ samples for diffusion-model post-training.

Preference-conditioned methods in multi-objective RL\citep{roijers2013survey,Hayes_2022} train policies to optimize user-specified reward weightings \citep{yang2019generalized,wang2024conditional}. In contrast, VPO learns to generate a set of solutions without conditioning on a particular weighting, using the R2 indicator \citep{brockhoff2012r2} with weighted-sum utilities as its objective. Related work has also documented the loss of output diversity during RL post-training \citep{kirk2024understanding,yue2025does}, motivating methods that preserve diversity while optimizing scalar rewards \citep{cui2025entropy,song2025outcome,li2025jointly,zhu2026flowrl}. App.~\ref{app:related_work} provides an extended discussion.

\section{Discussion and Conclusion}

We argued that when language models are deployed inside pipelines with test-time search, training should produce a diverse pool of competent candidates and leave exploitation to the search. VPO does this by sampling scalarizations uniformly over the simplex and using GRPO to train the policy to emit sets that span the supported trade-offs between reward components. Across MuSiQue, EUREQA, Maze, and ToolRL, VPO improves best@$k$ over scalar baselines, and the gap widens as the test-time budget grows.
We see two complementary explanations. First, the optima of $R(S)$ are reward-diverse by construction (Prop.~\ref{prop:optimal_sets}). Second, a candidate that scores poorly under $w^*$ but well under another $w$ still receives a positive gradient when that $w$ is sampled, while a fixed-$w^*$ run pushes it away. VPO may thus keep more strategies alive long enough to be refined, which could explain why it beats a $w^*$-trained policy even under $w^*$.

 VPO comes with four limitations. First, training compute is hard to equalize, since methods produce outputs of different lengths and VPO generates $m$ completions per rollout, but VPO still outperforms GRPO on MuSiQue when GRPO receives $3\times$ the rollouts (Sec.~\ref{sec:results}). Second, VPO benefits from a vector-valued reward and reduces to scalar best-of-$m$ training when the reward is scalar. Third, since VPO focuses on linear weightings $w^\top r$, it might not support optimizing nonlinear rewards at test time. Finally, VPO trades pass@$1$ for pass@$k$ and is meant for systems that include test-time search.

\begin{ack}
The authors express gratitude to Jyo Pari, Jonas H{\"u}botter, Nitish Dashora, Andre Ye, Itamar Pres, Navodita Sharma, John Marangola, Nolan Fey, Oliver Sieberling, Linlu Qiu, Luca Grillotti, Benjamin Eysenbach, Yoon Kim, Jacob Andreas, and members of the Improbable AI lab at MIT for discussions that helped shape this work.
This work was supported by National Science Foundation graduate research fellowships to RB, IP, and AK, the MIT-IBM Computing Research Lab, the MIT-Google Program for Computing Innovation and Hyundai Motor Company. This research was sponsored by the Army Research Office and was accomplished under Grant Numbers W911NF2110328 and W911NF-23-1-0277. The views and conclusions contained in this document are those of the authors and should not be interpreted as representing the official policies, either expressed or implied, of the Army Research Office or the U.S. Government. The U.S. Government is authorized to reproduce and distribute reprints for Government purposes notwithstanding any copyright notation herein.
\paragraph{Competing interests.}
The authors declare no competing interests.
\end{ack} 
\bibliography{references}

\appendix

\section{Theory of the Set-Level Objective}
\label{app:theory}

This appendix proves Proposition~\ref{prop:optimal_sets}. All statements concern the maximizers of $R$ for a single prompt. Remark~\ref{rem:scope} discusses what they do and do not imply about training.

\paragraph{Setup.}
Fix a prompt and drop it from the notation. Let $\mathcal{Y}$ be a finite set of candidate responses with reward vectors $r(y)\in\mathbb{R}^d$. Since $R(S)$ depends on $S$ only through the reward vectors of its elements, we treat responses with identical reward vectors as one response, so $r$ is one-to-one on $\mathcal{Y}$. A rollout outputs a set $S\subseteq\mathcal{Y}$ with $1\le|S|\le m$. Weightings lie in the simplex $\Delta=\{w\in\mathbb{R}^d_{\ge0}:\sum_i w_i=1\}$ and are drawn from $\mathrm{Dir}(\alpha)$ with $\alpha\in\mathbb{R}^d_{>0}$. We write
\[
R(S)=\mathbb{E}_{w}\Big[\max_{y\in S}w^\top r(y)\Big],\qquad
V^\star(w)=\max_{y\in\mathcal{Y}}w^\top r(y),\qquad
\mathrm{OPT}_m=\max_{|S|\le m}R(S).
\]

\begin{definition}[Supported response]
The \emph{cell} of $y\in\mathcal{Y}$ is the set of weightings under which $y$ is the unique best response,
\[
\Omega_y=\big\{w\in\Delta:\ w^\top r(y)>w^\top r(y')\ \text{for all } y'\neq y\big\}.
\]
A response is \emph{supported} if its cell is non-empty. We write $\mathcal{Y}_{\mathrm{s}}$ for the set of supported responses and $n_{\mathrm{s}}=|\mathcal{Y}_{\mathrm{s}}|$.
\end{definition}

In multi-objective RL, $\mathcal{Y}_{\mathrm{s}}$ is the smallest convex coverage set \citep{roijers2013survey}. We use one property of the Dirichlet distribution: \emph{every non-empty cell has positive probability}. A cell is defined by finitely many strict inequalities, so if it contains a weighting, it also contains all nearby weightings, including some with all components positive. The Dirichlet density is positive there.

\begin{lemma}[Supported responses are best under every weighting]\label{lem:cover}
For every $w\in\Delta$, some supported $y$ satisfies $w^\top r(y)=V^\star(w)$.
\end{lemma}
\begin{proof}
Call $w$ \emph{tied} if two responses share the highest score under $w$. For two distinct responses, the weightings with $w^\top r(y)=w^\top r(y')$ are either empty (if $r(y)-r(y')$ is a multiple of the all-ones vector) or a slice of $\Delta$ of lower dimension. A finite union of such slices contains no neighborhood, so every $w$ is the limit of untied weightings $w_n$. An untied $w_n$ lies in the cell of its best response $y_n$, so $y_n$ is supported. Since $\mathcal{Y}$ is finite, a single $y$ equals $y_n$ for infinitely many $n$, and by continuity $w^\top r(y)=\lim_n w_n^\top r(y)=\lim_n V^\star(w_n)=V^\star(w)$.
\end{proof}

\begin{lemma}[Every response is matched on average by supported ones]\label{lem:avg}
For every $y\in\mathcal{Y}$ there are weights $\lambda_z\ge0$, $z\in\mathcal{Y}_{\mathrm{s}}$, with $\sum_z\lambda_z=1$ and $r(y)\le\sum_{z}\lambda_z\,r(z)$ componentwise.
\end{lemma}
\begin{proof}
Consider the zero-sum game in which one player picks a mixture $\lambda$ of supported responses, the other picks $w\in\Delta$, and the payoff is $\phi(\lambda,w)=w^\top\big(\sum_z\lambda_z r(z)-r(y)\big)$. For fixed $w$, $\max_\lambda\phi(\lambda,w)=V^\star(w)-w^\top r(y)\ge0$ by Lemma~\ref{lem:cover}. As $\phi$ is bilinear and both strategy sets are simplices, the minimax theorem gives some $\lambda$ with $\phi(\lambda,w)\ge0$ for all $w$. Taking $w$ to be each vertex of $\Delta$ gives the claim.
\end{proof}

\begin{lemma}[Adding a missing supported response strictly helps]\label{lem:add}
If $y_0$ is supported and $y_0\notin S$, then $R(S\cup\{y_0\})>R(S)$ and $R(S)<\mathbb{E}_w[V^\star(w)]$.
\end{lemma}
\begin{proof}
For $w\in\Omega_{y_0}$, every $y\in S$ scores below $y_0$, so $\max_{y\in S}w^\top r(y)<w^\top r(y_0)=V^\star(w)$. Both differences in the claim are therefore non-negative everywhere and positive on $\Omega_{y_0}$, which has positive probability.
\end{proof}

\begin{proof}[Proof of Proposition~\ref{prop:optimal_sets}]
\emph{(i) $m\ge n_{\mathrm{s}}$.} For every $S$, $R(S)\le\mathbb{E}_w[V^\star(w)]$. If $S$ contains $\mathcal{Y}_{\mathrm{s}}$, Lemma~\ref{lem:cover} gives equality at every $w$, and such sets exist because $n_{\mathrm{s}}\le m$. So $\mathrm{OPT}_m=\mathbb{E}_w[V^\star(w)]$, and by Lemma~\ref{lem:add} a set that misses a supported response is not optimal. Hence every optimal set contains $\mathcal{Y}_{\mathrm{s}}$, and by Lemma~\ref{lem:cover} it contains a best response for every $w$, including $w^\star$.

\emph{(ii) $m<n_{\mathrm{s}}$.} Let $S^\star$ be optimal. (a) If $|S^\star|<m$, then $S^\star$ misses some supported $y_0$, and $S^\star\cup\{y_0\}$ has at most $m$ elements and is strictly better by Lemma~\ref{lem:add}, a contradiction. So $S^\star$ consists of $m$ distinct responses. (b) With $y_0\notin S^\star$ supported, $\mathrm{OPT}_{m+1}\ge R(S^\star\cup\{y_0\})>\mathrm{OPT}_m$. (c) Suppose $S^\star$ contains an unsupported $y$, let $T=S^\star\setminus\{y\}$ and $g(w)=\max_{y'\in T}w^\top r(y')$ (with $g=-\infty$ if $T$ is empty). Take $\lambda$ from Lemma~\ref{lem:avg} and let $Z$ be a supported response drawn with probabilities $\lambda$. For every $w$,
\[
\mathbb{E}_Z\big[\max\{g(w),\,w^\top r(Z)\}\big]\;\ge\;\max\big\{g(w),\,\mathbb{E}_Z[w^\top r(Z)]\big\}\;\ge\;\max\{g(w),\,w^\top r(y)\}.
\]
Averaging over $w$ gives $\sum_z\lambda_z R(T\cup\{z\})\ge R(S^\star)$, so $T\cup\{z\}$ is optimal for some supported $z$ and has one unsupported response fewer. Repeating this at most $m$ times gives an optimal set of supported responses only.
\end{proof}

\paragraph{Remarks.}
\begin{enumerate}[leftmargin=*,label=\arabic*.,ref=\arabic*]
\item\label{rem:unsupported} \textbf{Pareto optimality.} Supported responses are Pareto-optimal. If $r(y')\ge r(y)$ componentwise for some $y'\neq y$, then $w^\top r(y')\ge w^\top r(y)$ for every $w$, so $\Omega_y$ is empty. The converse, however, is not true. Next to $(1,0)$ and $(0,1)$, the response $(0.49,0.49)$ is Pareto-optimal but scores below $\max(w_1,w_2)$ under every weighting, so it is not supported. Every method that scores responses only through linear weightings $w^\top r$ shares this blind spot, including scalar RL with a fixed $w^\star$ and weight-conditioned multi-objective RL \citep{yang2019generalized}. A nonlinear downstream preference could, however, value such responses. Nonlinear scalarizations \citep{zhang2020random} would reward them, but the results above do not cover that case.

\item \textbf{Special cases.} For $m=1$, $R(\{y\})=\mathbb{E}[w]^\top r(y)$, which is scalar RL with the mean weighting. For $d=1$, $R(S)=\max_{y\in S}r(y)$ is the best-of-$m$ objective, and $n_{\mathrm{s}}=1$.

\item\label{rem:scope} \textbf{Scope.} The results describe the maximizers of $R$, not the training dynamics or what a policy of limited capacity learns. Because $\mathbb{E}_{S\sim\pi}[R(S)]$ is linear in the distribution of $S$, a policy maximizes it exactly when it puts all its probability on optimal sets. Parts (i), (ii-a) and (ii-b) need every non-empty cell to have positive probability, so they do not hold for scalar training with a fixed $w^\star$, whose optimum needs only one response. Training estimates $R$ with $K$ weightings that are redrawn at every step, which is unbiased for $R$.
\end{enumerate}

\section{Extended Related Work}\label{app:related_work}

This section extends Sec.~\ref{sec:related_work}. Works that appeared after May 2026 are concurrent with ours.

\paragraph{Training for inference-time selection.}
BOND and BoNBoN \citep{sessa2025bond,gui2024bonbon} distill the best-of-$N$ distribution into a single policy, and InfAlign \citep{balashankar2024infalign} aligns a policy for later best-of-$N$ decoding. Policy gradients for objectives over $k$ i.i.d.\ samples cover pass@$k$, max@$k$, and majority voting \citep{tang2025optimizing,walder2025pass,chen2025pass,chow_inference-aware_2025,zhang2025rspo,bagirov2025best}, and MaxRL \citep{tajwar2026maximum} learns from successful rollouts with a maximum-likelihood-style objective. Related work improves their baselines \citep{takashiro2026advantage}, generalizes them to order statistics \citep{parmas2026ordergrad}, shows that exploration emerges from max-over-retries objectives \citep{koyamada2022remax,nishimori2026emergence}, and extrapolates from small training budgets to large deployment budgets \citep{li2026whatshould}. Most of these methods optimize a scalar reward over independent samples. Mode conditioning instead allocates test-time samples across reasoning modes \citep{wu2026modeconditioning}. VPO generates the set in a single rollout and scores it under random scalarizations of a vector reward.

\paragraph{Set-level objectives under reward uncertainty}
ROSA \citep{gxchen2026usingrewarduncertaintyinduce}, optimizes a set function of rewards drawn from a distribution over reward functions, and notes that Eq.~\eqref{eq:reward} is the special case with the max set function and Dirichlet-distributed linear scalarizations. ReMax \citep{koyamada2022remax,nishimori2026emergence} studies expected-max objectives under reward uncertainty for exploration in RL. VPO shares this reward structure but generates candidates sequentially within one autoregressive rollout and assigns one advantage to the whole chain rather than one per action. Per-candidate credit, for example based on the scalarizations each candidate wins, is a natural extension. Concurrent work on text-to-image diffusion models, multi-axis max@$K$ \citep{onoda2026multiaxis}, takes the maximum score over a group of i.i.d.\ samples for each of a fixed set of target modes and sums these maxima. Up to a factor of $d$, this is Eq.~\eqref{eq:reward} with $w$ drawn uniformly from the vertices of the simplex rather than from a Dirichlet. Its credit assignment gives a sample weight on a mode only when it raises that mode's maximum, and it does not generate samples sequentially.
Multi-RLVR \citep{puri_reaching_2026} uses VPO's rollout format with a fixed scalar reward. Polychromic objectives \citep{hamid_polychromic_2025,orney2026poly}, DQO \citep{chen2026dqo}, SetPO \citep{li2026setpo}, GCPO \citep{chen2026gcpo}, and uniqueness-aware RL \citep{hu2026rewarding} encourage diverse generations through alternative set objectives or diversity measures such as embeddings, kernels, an LM judge, or task-specific distinctness tests. In VPO, a candidate's value is how much it raises the best scalarized task reward, so diversity is measured in task outcomes.

\paragraph{Multi-objective optimization and Quality-Diversity}

Preference-conditioned methods in multi-objective RL \citep{roijers2013survey,Hayes_2022,yang2019generalized,wang2024conditional} learn to optimize different reward weightings specified at inference time, while Rewarded Soups \citep{rame2023rewarded} interpolates models fine-tuned on different rewards. VPO instead generates a set of solutions without conditioning on a preference. Its objective is the R2 indicator with weighted-sum utilities \citep{brockhoff2012r2}, which rewards coverage of supported Pareto-optimal solutions but not necessarily the full Pareto front. Related ideas appear in lexicase selection \citep{spector_assessment_2012,helmuth_solving_2015,la_cava_probabilistic_2019}, which favors individuals under different orderings of test cases, and quality diversity methods \citep{bradley2024qdaif,samvelyan2024rainbow}, which maintain diverse solutions in an archive. VPO instead defines diversity through the task's reward components and trains a single policy to generate a set without an archive.

\paragraph{Diversity collapse.}
Post-training, including RLHF, reduces output diversity \citep{kirk2024understanding,karouzos2026does,gx2025kl} and can push pass@$k$ at large $k$ below that of the base model \citep{yue2025does,wu2025invisible}.
Proposed remedies control entropy \citep{cui2025entropy}, add outcome-level exploration bonuses \citep{song2025outcome}, preserve candidate-token support \citep{peng2025simko}, jointly reward quality and semantic diversity \citep{li2025jointly}, or match target distributions derived from the reward \citep{zhu2026flowrl,liu2026gflowrl,li2026dmpo}. These methods keep a scalar reward and change how probability is spread over responses. VPO changes what is rewarded, and a distribution-matching update could in principle be applied to its set-level reward.

\section{Environments}
\label{app:env_details}

For each domain we give the task and data, the reward vector $r(x,y)\in\mathbb{R}^d$, the target scalarization $w^\star$ (the weighting used as the GRPO reward and for best@$k$), the base model, and the evaluation set. Table~\ref{tab:per_domain_scalars} summarizes them. All reward components are non-negative, and higher is better. The prompts are listed in App.~\ref{app:prompts}, and training settings in App.~\ref{app:training_setup}. All base models are from the Qwen3 \citep{yang2025qwen3} and Qwen2.5-Coder \citep{hui2024qwen25coder} families.

\begin{table}[h]
\centering
\small
\setlength{\tabcolsep}{4pt}
\caption{Target scalarization $w^\star$ for each domain, with the reward dimension $d$, base model, and evaluation set. $\bar r$ is the uniform mean of the $d$ reward components.}
\label{tab:per_domain_scalars}
\begin{tabular}{l c l l l}
\toprule
Domain & $d$ & Target scalarization $w^{\star\top} r$ & Base model & Evaluation set \\
\midrule
Maze & $4$ & $\bar r$ & Qwen3-4B & 100 mazes \\
MuSiQue & $5$ & $\bigl(\sum_{i=1}^{4}\mathrm{hop}_i+3\,\mathrm{F1}\bigr)/7$ & Qwen3-1.7B & 300 questions \\
EUREQA & $5$ & $\bar r$ & Qwen3-8B & 682 questions \\
ToolRL & $4$ & $\bar r$ & Qwen3-1.7B & 80 prompts \\
LiveCodeBench & \#tests & $\bar r$ (pass fraction) & Qwen2.5-Coder-7B-Inst. & 279 problems \\
UltraFeedback & $5$ & overall score & Qwen3-4B & 200 prompts \\
\bottomrule
\end{tabular}
\end{table}

\subsection{Maze}
\label{app:env_maze}

\paragraph{Task and data.} The model outputs a sequence of moves (UP, DOWN, LEFT, RIGHT) that leads from a start corner $S$ to an exit corner $E$ of a $9{\times}9$ grid, collecting gold and diamond tiles and avoiding lava. Mazes are generated as follows.
(i)~Carve a spanning tree with randomized Prim's algorithm: start from a random cell and repeatedly open a random frontier wall cell that has exactly one open neighbor.
(ii)~Open $n_{\mathrm{cyc}}\sim\mathrm{Unif}\{18,\dots,28\}$ further wall cells that have at least two open neighbors, so that several routes exist.
(iii)~Place $S$ and $E$ on opposite corners of a random diagonal; the other two corners become the gold corner and the diamond corner.
(iv)~Let $L_G$, $L_D$, and $L_{GD}$ be the shortest-path lengths from $S$ to $E$ via the gold corner, via the diamond corner, and via both. Set the step budget to $T=\max(L_G,L_D)+7$ and reject the maze unless $L_{GD}>T$. A route within budget can therefore reach one item corner but not both.
(v)~Place $n_G,n_D\sim\mathrm{Unif}\{3,\dots,5\}$ gold and diamond tiles on open cells within Manhattan distance 2 of their corners, and $n_L\sim\mathrm{Unif}\{3,\dots,5\}$ lava tiles on open interior cells ($2\le \text{row},\text{col}\le 6$).
(vi)~Reject the maze unless a lava-free path from $S$ to $E$ of length at most $T$ exists.
The center cell is a bonus tile \texttt{B}. The prompt describes it as a score multiplier, but it has no effect on the reward; it is a distractor. We generate 1{,}000 training mazes and 100 evaluation mazes from disjoint fixed random seeds.

\paragraph{Reward vector.} The moves are read from the answer tags and simulated for at most $T$ steps. A move into a wall or off the grid leaves the agent in place but uses a step. Lava cells can be entered. The trajectory ends when the agent steps on $E$. If $E$ is not reached, $r=(0,0,0,0)$. Otherwise
\[
r(x,y) = (1,\ c_G,\ c_D,\ \ell),
\]
where $c_G$ and $c_D$ are the fractions of the gold and diamond tiles visited and $\ell$ is one minus the fraction of lava tiles visited. A tile counts once, and only if it is visited before $E$. The three item terms are therefore linear in the number of tiles and lie in $[0,1]$.

\paragraph{Target scalarization.} Uniform mean, $w^\star=\tfrac14\mathbf{1}$.

\paragraph{Model and evaluation.} Qwen3-4B, evaluated on the 100 held-out mazes.

\subsection{MuSiQue}
\label{app:env_musique}

\paragraph{Task and data.} MuSiQue-Ans \citep{trivedi2022musique} contains 2--4 hop questions, each composed of single-hop sub-questions. We train on the 19{,}938 training questions and evaluate on questions from the 2{,}417-question development split. Each prompt contains the question and the 20 paragraphs that MuSiQue provides with it (the 2--4 supporting paragraphs and distractors). The model lists supporting paragraph indices in \texttt{<support>} tags and gives its answer in \texttt{<answer>} tags.

\paragraph{Reward vector.} $r\in[0,1]^5$. For $i=1,\dots,4$, $\mathrm{hop}_i=1$ if the supporting paragraph of the $i$-th sub-question is cited. Only the first four distinct cited indices count, so citing every paragraph is not rewarded. For a question with $h<4$ hops, $\mathrm{hop}_i=1$ for $i>h$. The fifth component is the token-level F1 between the predicted answer and the gold answer (maximized over the gold answer's aliases) after lowercasing and removing punctuation and articles. A response without a parsable answer receives $r=\mathbf{0}$.

\paragraph{Target scalarization.} $w^{\star\top}r=(\mathrm{hop}_1+\mathrm{hop}_2+\mathrm{hop}_3+\mathrm{hop}_4+3\,\mathrm{F1})/7$, so the answer counts three times as much as any single citation.

\paragraph{Model and evaluation.} Qwen3-1.7B, trained for one epoch and evaluated on 300 development questions.

\subsection{EUREQA}
\label{app:env_eureqa}

\paragraph{Task and data.} Each EUREQA \citep{li2024deceptive} question describes a chain of six entities linked by five relations. The last entity is named in the text and the other five are masked (e.g.\ ``Person A'', ``Country B''). The model recovers the five masked entities by working back from the named one and writes each as an underscored Wikipedia title in tags \texttt{<entity\_A>} to \texttt{<entity\_E>}; entity A is the answer to the question. We use the five-hop questions, which come in a normal split (428 questions) and a hard split (1{,}363 questions with less identifying hints). We shuffle the hard split with a fixed seed and hold out half of it (682 questions) as the test set. The training set is the other half of the hard split plus the whole normal split (1{,}109 questions). The two splits are not disjoint at the level of entity chains: 156 of the 682 test questions share their entity chain with a training question, almost all of them normal-split questions with different wording and hints.

\paragraph{Reward vector.} $r\in\{0,1\}^5$, one exact-match indicator per masked entity A--E, after normalization (underscores to spaces, lowercasing, removing articles and punctuation, collapsing whitespace). A response in which all five entity tags are missing or empty receives $r=\mathbf{0}$.

\paragraph{Target scalarization.} Uniform mean.

\paragraph{Model and evaluation.} Qwen3-8B, trained for 6 epochs (102 steps) and evaluated on the 682 test questions; results are averaged over 4 evaluation seeds.

\subsection{ToolRL}
\label{app:env_toolrl}

\paragraph{Task and data.} We use the 4{,}000-example function-calling set released with ToolRL \citep{qian2025toolrl}, which combines examples from ToolACE \citep{liu2024toolace}, Hammer \citep{lin2024hammer}, and xLAM \citep{zhang2024xlam,liu2024apigen}, split into 3{,}920 training and 80 test prompts. The system message lists the available tools and the output format: a \texttt{<think>} block, optional JSON tool calls in \texttt{<tool\_call>}, and an optional \texttt{<response>}. The user message contains the dialogue. We use the prompts unchanged.

\paragraph{Reward vector.} $r\in[0,1]^4$.
\emph{Format} is 1 if the output has a \texttt{<think>} block and its tool-call section matches the structure of the gold answer (at least one parsable call if the gold answer has calls, no calls otherwise), and 0 otherwise.
\emph{Tool name} is the multiset F1 between predicted and gold tool names.
For the last two components, predicted calls are matched greedily to gold calls, first by name and then by overlap of parameter keys; unmatched calls on either side contribute 0.
\emph{Argument key} is the mean over matched pairs of the set F1 between predicted and gold parameter keys.
\emph{Argument value} is the mean over matched pairs of the average value score over the gold keys (token F1 for strings, exact match for numbers and booleans, multiset F1 for lists).
If neither the prediction nor the gold answer contains a call, the last three components are 1.

\paragraph{Target scalarization.} Uniform mean.

\paragraph{Model and evaluation.} Qwen3-1.7B, trained for 20 epochs (600 steps) and evaluated on the 80 test prompts; results are averaged over 4 evaluation seeds.

\subsection{LiveCodeBench}
\label{app:env_livecodebench}

\paragraph{Task and data.} We train on the DeepCoder corpus \citep{luo2025deepcoder}: 24{,}269 problems, of which 16{,}238 come from SYNTHETIC-1 \citep{primeintellect2025synthetic1}, 7{,}432 from the verified subset of TACO \citep{li2023taco}, and 599 from LiveCodeBench \citep{jain2024livecodebench} problems released between May 2023 and July 2024. Each problem has stdin/stdout or function-call test cases. We evaluate on the 279 LiveCodeBench problems released between August 2024 and February 2025, all later than every LiveCodeBench problem in the training data. This excludes overlap with the fine-tuning data, but not with the base model's pre-training data (Qwen2.5-Coder was released in September 2024).

\paragraph{Reward vector.} $r\in\{0,1\}^{d}$, one pass indicator per test case, where $d$ is the problem's number of test cases (at most 32 in training). VPO samples $w\sim\mathrm{Dir}(\mathbf{1}_d)$ with each problem's own $d$, without padding to a common length.

\paragraph{Target scalarization.} Uniform mean over the problem's test cases, i.e., the fraction of tests passed.

\paragraph{Model and evaluation.} Qwen2.5-Coder-7B-Instruct, trained for one epoch with GRPO and with VPO ($m=3$); the two runs differ in the prompt and the reward computation. For pass@$k$ and best@$k$, each method gets 30 candidates per problem on all 279 evaluation problems: 30 independent samples for GRPO and 10 chains of 3 for VPO, at temperature 0.8 and top-$p$ 0.95. For pass@$k$, a candidate is correct if it passes all test cases.

\paragraph{OpenEvolve.} We run OpenEvolve \citep{openevolve} with each trained model as the program generator. Each iteration produces 3 candidate programs: one 3-answer chain for VPO, and three independent samples for GRPO. A candidate's score is the fraction of the problem's tests it passes, and later iterations see this feedback. To select hard problems, we sample 30 candidates per problem from each model in a separate evaluation round with the same settings. We rank the 279 evaluation problems by the number of fully passing candidates (both models' 30 candidates combined) and then by the largest number of tests passed by any candidate, and take the 50 hardest. Neither model fully solves any of them in that round. Many candidates still pass some of the tests. We report the 32 of these for which both runs completed 200 iterations (about 600 candidates per problem and model). In the evaluation behind Fig.~\ref{fig:lcb_main_4panel}A,B, the mean best@$30$ on these 32 problems is $0.26$ for GRPO and $0.44$ for VPO.

\subsection{UltraFeedback}
\label{app:env_ultrafeedback}

\paragraph{Task and data.} Prompts come from UltraFeedback \citep{cui2023ultrafeedback}, and responses are scored by the ArmoRM reward model \citep{wang2024interpretable}.

\paragraph{Reward vector.} $r\in\mathbb{R}^5$: ArmoRM's five UltraFeedback objectives (overall score, instruction following, truthfulness, honesty, helpfulness).

\paragraph{Target scalarization.} Evaluation uses the overall-score component, $w^\star=e_1$.

\paragraph{Model and evaluation.} Qwen3-4B, trained for one epoch with GRPO, GDPO, goal-conditioned GRPO, Multi-RLVR, and VPO; the multi-answer methods use $m=3$ and responses of up to 3{,}072 tokens. We evaluate on 200 held-out prompts, with 50 samples per prompt for single-answer methods and 17 chains of 3 (51 candidates) for multi-answer methods.

\section{Training and Evaluation Protocol}
\label{app:training_setup}

\subsection{Shared recipe}
\label{app:shared_recipe}

All methods are implemented in veRL \citep{sheng2024hybridflow}, with vLLM \citep{kwon2023vllm} for generation, and share the settings in Table~\ref{tab:shared_hparams}. Methods differ in the prompt (single- or multi-answer), in how a rollout's reward is computed, and in the advantage estimator (App.~\ref{app:method_settings}). We train one run per method and domain with fixed random seeds and evaluate the final checkpoint. Table~\ref{tab:per_domain_hparams} lists the per-domain settings.

\begin{table}[h]
\centering
\small
\caption{Settings shared by all methods and domains.}
\label{tab:shared_hparams}
\begin{tabular}{l p{0.68\linewidth}}
\toprule
Policy loss & PPO clipped surrogate \citep{schulman2017proximal}, $\epsilon=0.2$, dual-clip constant 3; loss averaged over tokens \\
Updates per batch & one pass over the batch in 2 mini-batches (2 optimizer steps) \\
Advantage & one scalar per rollout, applied to all of its response tokens; no value network \\
KL regularization & loss term with coefficient $10^{-3}$, $k_3$ estimator \citep{schulman2020klapprox}, reference = initial checkpoint; no KL penalty in the reward \\
Entropy bonus & none \\
Optimizer & AdamW, learning rate $10^{-6}$ (constant, no warm-up), $(\beta_1,\beta_2)=(0.9,0.999)$, weight decay 0.01, gradient-norm clipping at 1.0 \\
Rollouts per prompt & $n=8$ \\
Training sampler & temperature 1.0, top-$p$ 1.0, no top-$k$ truncation \\
Chat template & Qwen3 thinking mode disabled \\
Precision and sharding & bf16 mixed precision, fully sharded data parallelism, gradient checkpointing \\
Hardware & one node with $4\times$H100 80\,GB per run ($8\times$H100 for LiveCodeBench) \\
\bottomrule
\end{tabular}
\end{table}

\begin{table}[h]
\centering
\footnotesize
\setlength{\tabcolsep}{3pt}
\caption{Per-domain settings. Batch is the number of prompts per training step. Lengths are the maximum prompt/response tokens for single-answer and for multi-answer methods; goal-conditioned GRPO uses the multi-answer prompt length and the single-answer response length. The evaluation pool is the number of candidates per prompt ($\times4$: four evaluation seeds).}
\label{tab:per_domain_hparams}
\begin{tabular}{l r c c c c c c}
\toprule
Domain & Train size & Epochs (steps) & Batch & Single-answer & Multi-answer & Eval.\ $T$ / top-$p$ & Eval.\ pool \\
\midrule
Maze & 1{,}000 & 3 & 128 & 512 / 512 & 768 / 1024 & 0.7 / 1.0 & 30 \\
MuSiQue & 19{,}938 & 1 (157) & 128 & 4096 / 512 & 5120 / 1024 & 0.7 / 1.0 & 30 \\
EUREQA & 1{,}109 & 6 (102) & 64 & 1024 / 512 & 1024 / 2048 & 0.7 / 0.95 & 30$\times$4 \\
ToolRL & 3{,}920 & 20 (600) & 128 & 4096 / 1024 & 4352 / 2048 & 0.7 / 0.95 & 30$\times$4 \\
LiveCodeBench & 24{,}269 & 1 (378) & 64 & 4096 / 4096 & 4096 / 4096 & 0.8 / 0.95 & 30 \\
UltraFeedback & -- & 1 & -- & -- & -- / 3072 & -- & 50 / 51 \\
\bottomrule
\end{tabular}
\end{table}

\subsection{Method-specific settings}
\label{app:method_settings}

Let $s_i$ be the scalar reward of rollout $i$ in a group of $n$ rollouts for the same prompt, and $\mu$ and $\sigma$ the group mean and standard deviation of $s$.

\paragraph{GRPO} \citep{shao2024deepseekmath}. Single-answer prompt, $s_i=w^{\star\top}r_i$, and $A_i=(s_i-\mu)/(\sigma+10^{-6})$, with $\sigma$ the sample standard deviation of the group.

\paragraph{Random-$w$.} As GRPO, but each training example has its own weighting $w\sim\mathrm{Dir}(\mathbf{1})$, drawn once and kept for all epochs, and $s_i=w^\top r_i$. The weighting is not shown to the model.

\paragraph{Goal-conditioned GRPO.} As Random-$w$, but the weighting is appended to the prompt (App.~\ref{app:prompts_goalcond}). At evaluation the prompt contains either $w^\star$ or a fresh $w\sim\mathrm{Dir}(\mathbf{1})$ for each sample. We ran this baseline on Maze (Table~\ref{tab:maze-goalcond}) and UltraFeedback (Table~\ref{tab:ultrafeedback}).

\paragraph{Max@$k$} \citep{bagirov2025best}. Single-answer prompt. Each rollout's advantage is its weight in the unbiased estimator of the expected maximum of $s$ over a random $k$-subset of the group, followed by group mean and standard-deviation normalization. We use $k=4$.

\paragraph{MaxRL} \citep{tajwar2026maximum}. Single-answer prompt and $A_i=(s_i-\mu)/(\mu+10^{-6})$, with the same group size as GRPO.

\paragraph{Multi-RLVR} \citep{puri_reaching_2026}. Multi-answer prompt with $m=3$. A rollout's reward is the sum of $w^{\star\top}r$ over its $m$ answer slots after parsing and de-duplication (App.~\ref{app:parsing}); the advantage is computed as in GRPO.

\paragraph{GDPO} \citep{liu_gdpo_2026}. Single-answer prompt. Each reward component is normalized within the group as in GRPO, the $d$ normalized values are summed with equal weights, and the sum is standardized across the batch. Used in the rollout-matched ablation (Table~\ref{tab:musique-3x-ablation}) and on UltraFeedback.

\paragraph{VPO.} Multi-answer prompt with $m=3$. For each prompt we draw $K=100$ weightings $w^{(1)},\dots,w^{(K)}\sim\mathrm{Dir}(\mathbf{1})$, shared by the $n$ rollouts of the group. Rollout $g$ with answer set $S^{(g)}$ receives
\[
\hat R(S^{(g)})=\frac1K\sum_{k=1}^K \max_{y\in S^{(g)}} w^{(k)\top} r(x,y)
\]
and the advantage $(\hat R(S^{(g)})-\mu)/(\sigma+10^{-6})$, with $\sigma$ the population standard deviation of the group. If $\sigma<10^{-6}$, all advantages in the group are zero.

\paragraph{Rollout-matched ablation.} GRPO, GDPO, and Random-$w$ with $n=24$ rollouts per prompt and otherwise unchanged settings.

\subsection{Parsing multi-answer outputs}
\label{app:parsing}

Multi-RLVR and VPO read the $m$ numbered answer blocks requested by the prompt (\texttt{<route\_i>} for Maze, \texttt{<response\_i>} otherwise).
\begin{itemize}[leftmargin=*,itemsep=1pt,topsep=2pt]
\item Only blocks $1,\dots,m$ are read; further answers are ignored.
\item A missing or unparsable block receives the zero reward vector (Maze: no moves; MuSiQue: no answer; EUREQA: all entity tags empty; ToolRL: empty block). A rollout without any parsable answer therefore has $\hat R=0$.
\item During training, duplicate answers are removed before the rollout reward is computed, keeping the first occurrence. Duplicates are answers with identical reward vectors (MuSiQue, EUREQA, ToolRL) or identical move sequences (Maze). Removed answers are replaced by zero vectors, so each rollout keeps $m$ rows. This does not change VPO's reward, which is a maximum over non-negative scores, but it removes the credit that Multi-RLVR's sum would give to repeated answers.
\item At evaluation nothing is removed: every slot is scored as generated, and missing slots count as zero vectors.
\end{itemize}

\subsection{Evaluation metrics}
\label{app:metrics}

For every evaluation prompt we draw a pool of $N=30$ candidates: 30 independent samples for single-answer methods and 10 chains of $m=3$ answers for multi-answer methods (ToolRL: the first 30 of 48 samples; UltraFeedback: App.~\ref{app:env_ultrafeedback}). Each candidate $y$ is scored by $v=w^{\star\top}r(x,y)$.

\paragraph{best@$k$ (tables).} Let $v_{(1)}\ge\dots\ge v_{(N)}$ be the sorted scores of a prompt's pool. We report
\[
\mathrm{best}@k=\sum_{i=1}^{N-k+1}\frac{\binom{N-i}{k-1}}{\binom{N}{k}}\,v_{(i)},
\]
the expected maximum over a uniformly random $k$-subset of the pool, averaged over prompts (and over evaluation seeds for EUREQA and ToolRL). For multi-answer methods the subset is drawn from all 30 candidates, regardless of chain. Table~\ref{tab:maze-goalcond} uses the same estimator.

\begin{figure}[h]
    \centering
    \includegraphics[width=\linewidth]{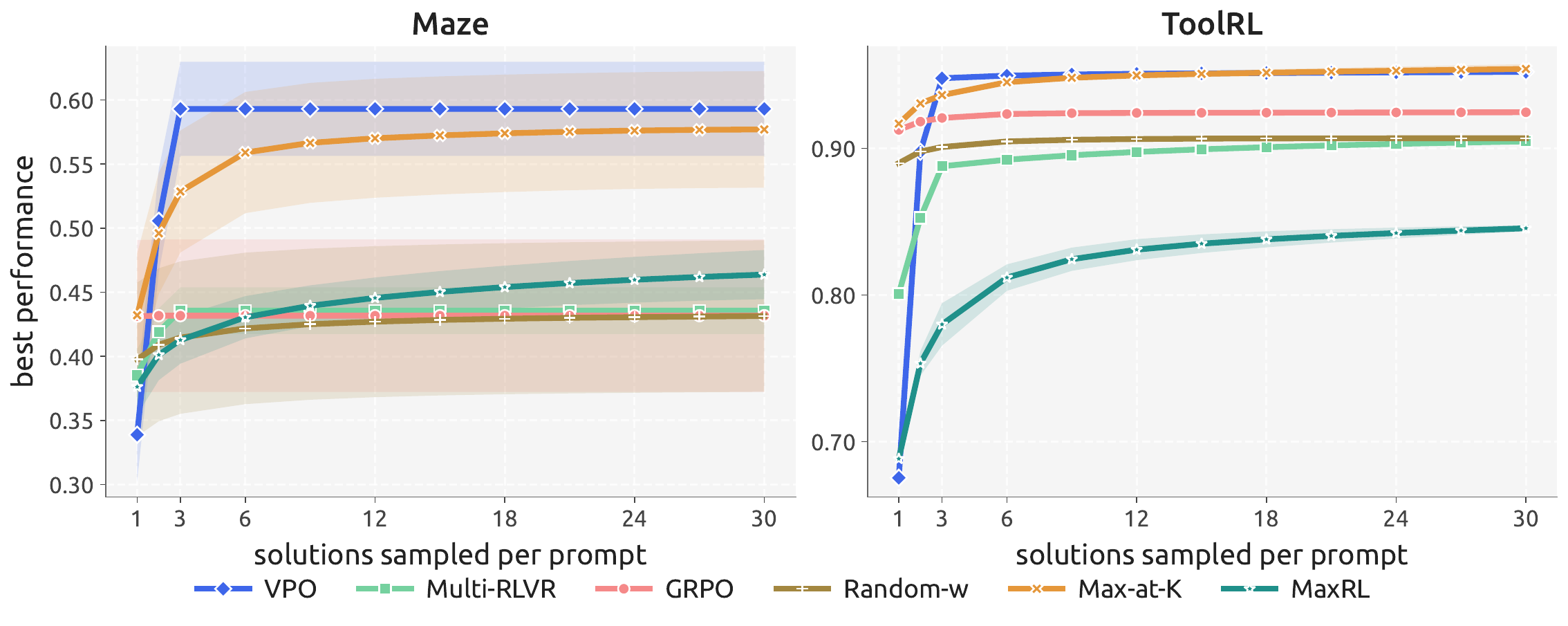}
    \caption{\textbf{Test-time scaling on Maze and ToolRL.} Chain-pooled best@$k$ under $w^\star$ (App.~\ref{app:metrics}) as a function of the number of candidates $k$; companion to Fig.~\ref{fig:chain_scaling}. Shaded bands are 95\% confidence intervals across mazes (Maze) and across 4 evaluation seeds (ToolRL). On Maze, Max@$k$ and GRPO are highest at $k=1$ and VPO is highest from $k=2$ on. On ToolRL, Max@$k$ is highest for $k\le 2$; from $k=3$ on, VPO and Max@$k$ are above the other methods and differ by at most 0.011.}
    \label{fig:chain_scaling_maze_toolrl}
\end{figure}

\paragraph{Chain-pooled best@$k$ (Figs.~\ref{fig:chain_scaling} and~\ref{fig:chain_scaling_maze_toolrl}).} The 30 candidates are grouped into 10 chains of 3 (for single-answer methods, consecutive independent samples). For $k\le 3$, best@$k$ is the expected maximum over a random $k$-subset of one chain, averaged over chains. For $k=3c$, it is the expected maximum over $c$ chains drawn without replacement, where each chain contributes its best candidate.

\paragraph{$\mathbb{E}_w[\mathrm{best}@k]$.} best@$k$ with candidates scored by $w^\top r$ instead of $w^{\star\top} r$, averaged over $w\sim\mathrm{Dir}(\mathbf{1})$ (200 draws for VPO in Table~\ref{tab:maze-goalcond}).

\paragraph{pass@$k$ (LiveCodeBench).} The unbiased estimator of \citet{chen2021evaluating} over the 30-candidate pool, where a candidate is correct if it passes all test cases. best@$k$ on LiveCodeBench scores a candidate by the fraction of tests it passes.

\paragraph{F1@30 (MuSiQue).} The largest answer F1 among the 30 candidates, averaged over questions.

\paragraph{Reward-space diversity (tables).} For each prompt, the mean of $\lVert r(x,y_i)-r(x,y_j)\rVert_1$ over all $\binom{30}{2}=435$ pairs of candidates in the evaluation pool (chains pooled), averaged over prompts and seeds. Unparsable candidates enter as zero vectors. The largest possible value is $d$.

\begin{figure}[h]
    \centering
    \includegraphics[width=\linewidth]{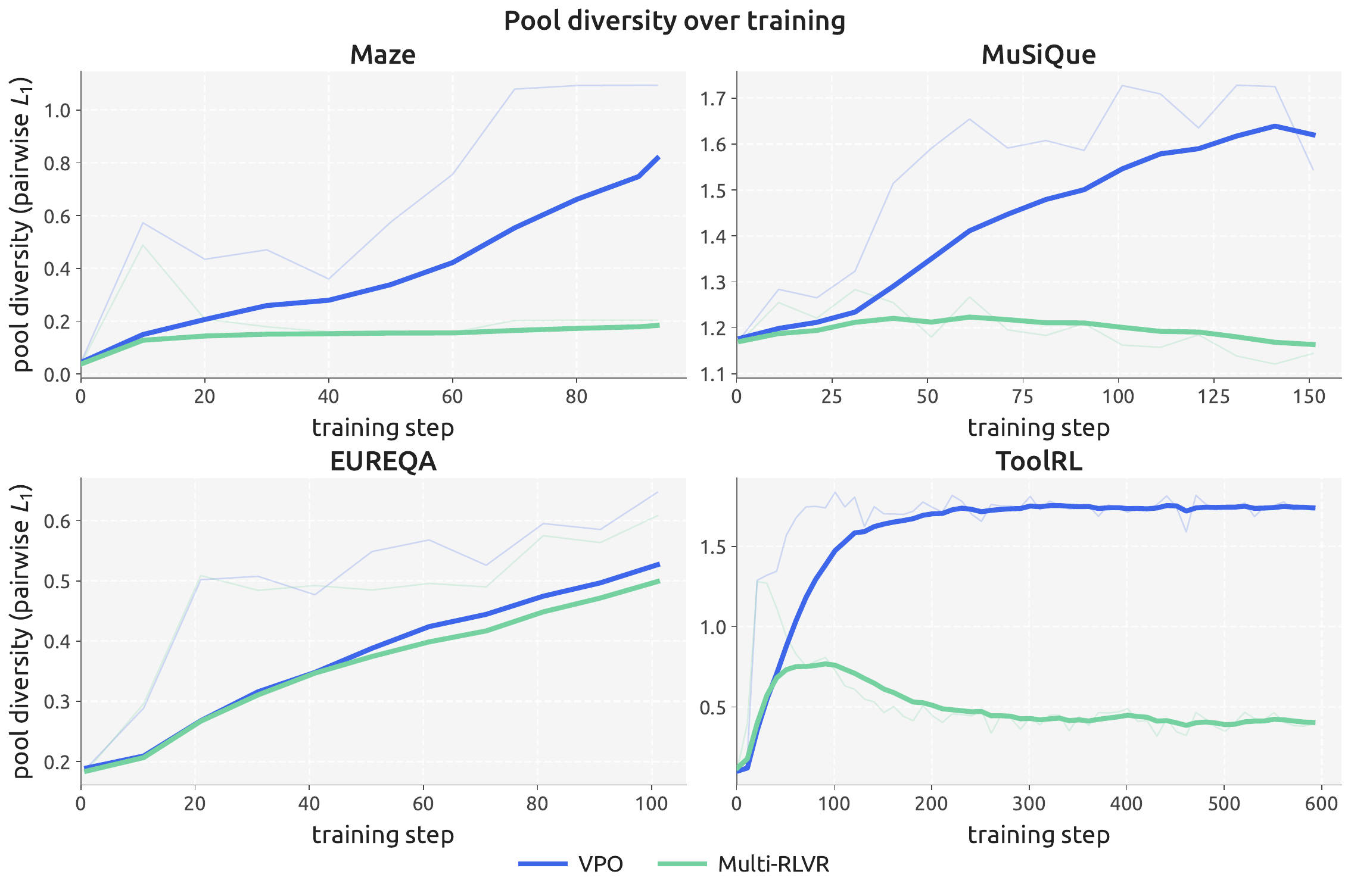}
    \caption{\textbf{Reward-space diversity during training} for VPO and Multi-RLVR: the mean pairwise $L_1$ distance between the reward vectors of the 24 candidates in each prompt's rollout group, averaged over the training batch (App.~\ref{app:metrics}). Faint lines are the logged values, subsampled to a similar number of points per panel; solid lines are exponential moving averages (smoothing factor 0.2). Each panel ends at the last step of the shorter run. This is a training-time statistic and is not the evaluation diversity reported in the tables. VPO's diversity is higher than Multi-RLVR's on Maze, MuSiQue, and ToolRL; on EUREQA the two are close.}
    \label{fig:diversity_over_time}
\end{figure}

\paragraph{Training-time diversity (Fig.~\ref{fig:diversity_over_time}).} The same statistic computed at each training step over the $n\cdot m=24$ candidates of each prompt's rollout group and averaged over the prompts in the batch. It differs from the evaluation statistic in two ways: it uses training samples (temperature 1.0), and removed duplicates and unparsable answers enter as the zero vectors substituted during training.

\section{Additional Analyses}
\label{app:additional_analyses}

\subsection{Additional figures}
\label{app:additional_figures}

\subsection{Rollout-matched ablation}
\label{app:rollout_ablation}

Table~\ref{tab:musique-3x-ablation} (Sec.~\ref{sec:results}) compares VPO at $n=8$ with GRPO, GDPO, and Random-$w$ at $n=24$ on MuSiQue. With $n=24$ the baselines receive as many reward evaluations per prompt as VPO ($8$ rollouts $\times$ $3$ answers) and three times as many generated rollouts. All other settings follow App.~\ref{app:training_setup}.

\subsection{Reward collinearity and when VPO helps}
\label{app:ultrafeedback}

We examine the relationship between reward-component correlation and VPO's improvement over GRPO. Table~\ref{tab:rho-vs-bestk} reports the mean pairwise Pearson correlation $\bar\rho$ between reward components, measured on samples from the trained models, together with best@16. These analyses use the same evaluation pools and target scoring rules as the corresponding experimental tables. Because $\bar\rho$ is measured after training, it describes the trained policies' outputs and is not a prediction made before training.

On Maze, MuSiQue, and EUREQA, $\bar\rho\le 0.39$ for both methods and VPO has the higher best@16. ToolRL does not follow a simple threshold: $\bar\rho_{\mathrm{VPO}}=0.86$, yet VPO is still ahead by 0.027. On UltraFeedback the components are strongly correlated for both methods ($\bar\rho=0.95$ for VPO and $0.82$ for GRPO), and VPO roughly matches GRPO (Table~\ref{tab:ultrafeedback}): it has the lowest best@1 among trained methods but is within 0.005 of GRPO from best@10 on. With five domains, the association between high $\bar\rho$ and a small gain is an observation, not a tested relationship.

\begin{table}[h]
\centering
\small
\caption{\textbf{Best@$k$ on UltraFeedback} (ArmoRM reward, $w^\star$ = overall score, 200 held-out prompts). $\Delta=\text{best@}50-\text{best@}1$. Bold and underline mark the best and second-best trained method in each column (base-model rows excluded). No multi-answer base-model evaluation was run; its entries are shown as ``--''.}
\label{tab:ultrafeedback}
\begin{tabular}{l c c c c c c}
\toprule
Method & best@$1$ & best@$5$ & best@$10$ & best@$20$ & best@$50$ & $\Delta$ \\
\midrule
Qwen3-4B (single-answer prompt) & $0.700$ & $0.724$ & $0.730$ & $0.735$ & $0.741$ & $+0.041$ \\
\hspace{1em}$\hookrightarrow$ GRPO & $\mathbf{0.741}$ & $\underline{0.762}$ & $\underline{0.768}$ & $\underline{0.773}$ & $\underline{0.779}$ & $+0.038$ \\
\hspace{1em}$\hookrightarrow$ GDPO & $\underline{0.737}$ & $0.759$ & $0.765$ & $0.770$ & $0.776$ & $+0.039$ \\
\hspace{1em}$\hookrightarrow$ Goal-cond.\ GRPO & $0.727$ & $0.750$ & $0.756$ & $0.762$ & $0.767$ & $+0.040$ \\
\midrule
Qwen3-4B (multi-answer prompt) & -- & -- & -- & -- & -- & -- \\
\hspace{1em}$\hookrightarrow$ Multi-RLVR & $0.730$ & $\mathbf{0.763}$ & $\mathbf{0.770}$ & $\mathbf{0.775}$ & $\mathbf{0.780}$ & $+0.050$ \\
\hspace{1em}$\hookrightarrow$ VPO (ours) & $0.700$ & $0.755$ & $0.763$ & $0.769$ & $0.776$ & $+0.076$ \\
\bottomrule
\end{tabular}

\end{table}

\begin{table}[h]
\centering
\small
\caption{\textbf{Reward collinearity and best@16.} $\bar\rho$ is the mean off-diagonal Pearson correlation between reward components over all samples from the trained model; $\Delta$ is VPO minus GRPO best@16, computed before rounding. Bold marks the higher best@16.}
\label{tab:rho-vs-bestk}
\begin{tabular}{l c c c c c c}
\toprule
 & & \multicolumn{2}{c}{$\bar\rho$} & \multicolumn{2}{c}{best@$16$} & \\
\cmidrule(lr){3-4} \cmidrule(lr){5-6}
Domain & $d$ & VPO & GRPO & VPO & GRPO & $\Delta$ \\
\midrule
Maze & $4$ & $0.39$ & $0.37$ & $\mathbf{0.593}$ & $0.432$ & $+0.161$ \\
MuSiQue & $5$ & $0.13$ & $0.15$ & $\mathbf{0.820}$ & $0.724$ & $+0.096$ \\
EUREQA & $5$ & $0.03$ & $0.03$ & $\mathbf{0.267}$ & $0.231$ & $+0.036$ \\
ToolRL & $4$ & $0.86$ & $0.62$ & $\mathbf{0.951}$ & $0.924$ & $+0.027$ \\
UltraFeedback & $5$ & $0.95$ & $0.82$ & $0.767$ & $\mathbf{0.772}$ & $-0.004$ \\
\bottomrule
\end{tabular}

\end{table}

\paragraph{Method.} For each domain and method, we use the saved evaluation samples underlying the corresponding experimental table.
\begin{enumerate}[leftmargin=*,itemsep=1pt,topsep=2pt]
\item \textbf{Sample pool.} The pools (prompts $\times$ candidates) are: Maze $100\times30$, MuSiQue $300\times30$, EUREQA $682\times30$ for each of 4 evaluation seeds, ToolRL $80\times30$ for each of 4 evaluation seeds, and UltraFeedback $200\times51$ (VPO) or $200\times50$ (GRPO).
\item \textbf{$\bar\rho$.} Stack the reward vectors across prompts, candidates, and evaluation seeds into a matrix with $d$ columns, drop components with zero variance, compute the Pearson correlation matrix, and average its off-diagonal entries.
\item \textbf{best@16.} Score candidates using the domain's target scalarization $w^\star$ and apply the random-subset estimator of App.~\ref{app:metrics} with $k=16$, averaging over prompts and, where applicable, evaluation seeds.
\end{enumerate}

\section{Extra Experimental Results}
\label{app:extra_results}

This appendix reports three additional experiments. They measure how sensitive VPO is to the number of sampled weightings, whether a higher sampling temperature closes the gap between GRPO and VPO, and how many distinct candidates each method produces.

\subsection{Number of weight draws}
\label{app:weight_draws}

VPO estimates $R(S)$ with $K$ weightings drawn per prompt group (\S\ref{sec:vpo_objective}), and all main experiments use $K=100$. To measure the sensitivity to $K$, we retrain VPO on MuSiQue from scratch with $K\in\{1,4,16,100\}$, changing nothing else, and evaluate best@$k$ under $w^\star$ (Table~\ref{tab:weight_draws}). These runs are separate from those in Table~\ref{tab:musique}, so the absolute values differ slightly. With a single draw, this produces a noisy, potentially unstable learning signal, and best@$30$ drops to $0.688$. From $K=4$ on, the results are flat, and the remaining differences are within run-to-run variation.

\begin{table}[h]
\centering
\small
\caption{\textbf{Number of weight draws $K$} (MuSiQue, best@$k$ under $w^\star$). Separate training runs from Table~\ref{tab:musique}.}
\label{tab:weight_draws}
\begin{tabular}{c c c c}
\toprule
$K$ & best@$1$ & best@$3$ & best@$30$ \\
\midrule
$1$ & $0.529$ & $0.622$ & $0.688$ \\
$4$ & $0.666$ & $0.769$ & $0.864$ \\
$16$ & $0.682$ & $0.775$ & $0.862$ \\
$100$ & $0.660$ & $0.764$ & $0.860$ \\
\bottomrule
\end{tabular}
\end{table}

Sampling more weightings adds negligible cost. Each candidate is scored once and its reward vector is cached, so the $K$ Monte Carlo terms reduce to dot products. On MuSiQue with Qwen3-1.7B, going from $K=1$ to $K=100$ raises the advantage computation from $0.031$ to $0.043$ seconds per step, about $0.01\%$ of a training step of roughly $117$ seconds.

\subsection{Sampling temperature}
\label{app:temperature}

A higher sampling temperature also spreads a policy's outputs, so we test whether it closes the gap between GRPO and VPO. We train one GRPO and one VPO policy each on MuSiQue and EUREQA (separate runs from the main tables), sample $30$ candidates per prompt at temperatures from $0.7$ to $1.5$, and report best@$30$ under $w^\star$ (Table~\ref{tab:temperature}). On MuSiQue, a higher temperature improves both methods by about $0.02$, and the gap stays near $0.08$. GRPO at its best temperature remains $0.059$ below VPO at its worst, and best@$1$ does not change with temperature for either method (GRPO $0.735$ to $0.733$, VPO $0.663$ to $0.660$). On EUREQA, a higher temperature narrows the gap from $0.034$ to $0.014$, and VPO stays ahead at every temperature. How much temperature helps a scalar-trained policy therefore depends on the domain, but it does not close the gap on either domain.

\begin{table}[h]
\centering
\small
\caption{\textbf{Sampling temperature} (best@$30$ under $w^\star$). Separate training runs from the main tables. $\pm$ is one standard deviation.}
\label{tab:temperature}
\begin{tabular}{l l c c c c}
\toprule
Domain & Method & $T=0.7$ & $T=1.0$ & $T=1.2$ & $T=1.5$ \\
\midrule
MuSiQue & GRPO & $0.767\pm.002$ & $0.774\pm.004$ & $0.781\pm.005$ & $0.788\pm.005$ \\
 & VPO (ours) & $\mathbf{0.847\pm.005}$ & $\mathbf{0.851\pm.006}$ & $\mathbf{0.859\pm.005}$ & $\mathbf{0.865\pm.005}$ \\
 & gap & $+0.080$ & $+0.077$ & $+0.078$ & $+0.077$ \\
\midrule
EUREQA & GRPO & $0.251\pm.005$ & $0.262\pm.007$ & $0.270\pm.008$ & $0.273\pm.009$ \\
 & VPO (ours) & $\mathbf{0.285\pm.017}$ & $\mathbf{0.291\pm.018}$ & $\mathbf{0.289\pm.020}$ & $\mathbf{0.287\pm.018}$ \\
 & gap & $+0.034$ & $+0.028$ & $+0.019$ & $+0.014$ \\
\bottomrule
\end{tabular}
\end{table}

\subsection{Distinct candidates}
\label{app:distinct_candidates}

Table~\ref{tab:distinct} counts distinct candidates in the MuSiQue evaluation pools behind Table~\ref{tab:musique} for the base model, GRPO, and VPO. Two candidates are distinct when their reward vectors differ, and a distinct candidate counts as competent when its score under $w^\star$ is at least half the maximum possible score. For VPO, a group of three is one chain of three candidates. For the base model and GRPO, which produce one answer per sample, a group is three independent samples. Scalar RL collapses the pool. The base model, sampled at temperature $0.7$ without top-$k$ or top-$p$ truncation, already produces more than one distinct reward vector in $45.4\%$ of its groups of three candidates, and GRPO lowers this to $10.8\%$. Since the decoding settings are the same, the collapse comes from training. VPO raises the fraction to $85.3\%$. The base model has the most distinct candidates per pool but the lowest best@$30$, because many of its candidates score zero on every component. VPO keeps most of the base model's distinct candidates ($3.33$ of $4.04$ per pool) and makes far more of them competent ($2.44$ against $0.88$).

\begin{table}[h]
\centering
\small
\caption{\textbf{Distinct candidates on MuSiQue} (pools of $30$ candidates per prompt, from the evaluations behind Table~\ref{tab:musique}). Candidates are distinct when their reward vectors differ.}
\label{tab:distinct}
\begin{tabular}{l c c c}
\toprule
 & Base model & GRPO & VPO (ours) \\
\midrule
Distinct candidates per pool & $\mathbf{4.04}$ & $1.33$ & $3.33$ \\
Groups of three with $>1$ distinct candidate & $45.4\%$ & $10.8\%$ & $\mathbf{85.3\%}$ \\
Competent distinct candidates per pool & $0.88$ & $1.09$ & $\mathbf{2.44}$ \\
best@$30$ & $0.423$ & $0.728$ & $\mathbf{0.832}$ \\
\bottomrule
\end{tabular}
\end{table}

\section{Compute Resources}
\label{app:compute}

All experiments ran on NVIDIA H100 80\,GB GPUs. Table~\ref{tab:compute} lists the runs behind the reported results, including evaluation (sampling the candidate pools and computing rewards) and the OpenEvolve runs. A GPU-hour is wall-clock time times the number of GPUs, summed over restarts of the same run. Most values come from our training logs. Rows marked est.\ had no log and are extrapolated from comparable runs. The reported experiments total roughly $1{,}850$ GPU-hours. Including preliminary and failed runs, the full project used roughly $4{,}500$ GPU-hours.

\begin{table}[h]
\centering
\small
\caption{Runs and compute. GPU-hours are approximate, and est.\ marks rows extrapolated from comparable runs.}
\label{tab:compute}
\begin{tabular}{p{0.46\linewidth} c c r}
\toprule
Experiment & Runs & GPUs per run & GPU-hours \\
\midrule
Main grid: 4 domains $\times$ \{GRPO, Random-$w$, Max@$k$, MaxRL, Multi-RLVR, VPO\} & 24 & 4--8$\times$H100 & $\approx$350 (part est.) \\
Goal-conditioned GRPO (Maze) & 1 & $4\times$H100 & $\approx$8 (est.) \\
GDPO, $n=8$ (MuSiQue) & 1 & $4\times$H100 & $\approx$11 \\
Rollout-matched ablation, $n=24$ (MuSiQue) & 3 & $4\times$H100 & $\approx$100 \\
UltraFeedback (5 methods) & 5 & 4--8$\times$H100 & $\approx$335 \\
LiveCodeBench (GRPO, VPO) & 2 & $8\times$H100 & $\approx$785 \\
OpenEvolve (2 models, 50 problems, 200 iterations) & 8 shards & 1 GPU per shard & $\approx$64 \\
Extra experiments (App.~\ref{app:extra_results}): weight draws, temperature & 8 & $4\times$H100 & $\approx$90 (est.) \\
Evaluation (all pools) & -- & 1 GPU & $\approx$100 (est.) \\
\midrule
Total & & & $\approx$1{,}850 \\
\bottomrule
\end{tabular}
\end{table}

\FloatBarrier
\section{Domain Prompts}
\label{app:prompts}

Below are the prompt templates used in training. For each domain we show the single-answer prompt and then the change made for the multi-answer methods (Multi-RLVR and VPO), which use $m=3$. Text in braces is filled in per example. Em dashes in the original templates are rendered as \texttt{--}. Goal-conditioned GRPO appends the suffix in App.~\ref{app:prompts_goalcond} to the single-answer prompt.

\subsection{Maze}
\label{app:prompts_maze}

The example shows the prompt for one maze; the grid, item counts, and step budget are filled in per maze.

\begin{promptbox}{Maze: single-answer prompt}
Navigate a 9x9 maze from S to E. Collect gold and diamonds, avoid lava.

Grid:
S . . . . . . D .
# . # . . . . D D
. . # # . . . . D
. # L . . . L # .
. . . . B . . . .
. . . . . L . . .
G . . . . . . . .
G G . . . . . . .
G G . . . . # . E

- Move: UP, DOWN, LEFT, RIGHT. # is a wall -- you cannot enter it.
- Do not leave the grid.
- Collect: G (Gold), D (Diamond), B (Bonus) tiles by stepping on them.
- Avoid: L (Lava) tiles. Stepping on lava costs you.
- Visiting a B cell multiplies your other scores -- explore!
- You MUST reach E. If you don't reach E, your score is zero everywhere.
- Items only count if collected BEFORE you reach E (the trajectory ends at E).
- You have 27 steps.

This maze has 5 Gold, 4 Diamond, 3 Lava, and 1 Bonus tiles.
Output moves in <answer>...</answer> tags, e.g., <answer>UP UP RIGHT</answer>.
\end{promptbox}

\paragraph{Multi-answer rewrite.} ``You have 27 steps.'' becomes ``You have 27 steps per route.'', and the last line (``Output moves \ldots'') is replaced by the block below.

\begin{promptbox}{Maze: multi-answer rewrite}
Reason briefly about the maze, then provide 3 genuinely different routes from S to E. Each route is a sequence of UP/DOWN/LEFT/RIGHT moves (space-separated). Wrap each route in numbered tags (<route_1>...</route_1>, <route_2>...</route_2>, <route_3>...</route_3>). Inside each tag put ONLY moves (no arrows, no coordinates, no prose); any reasoning goes outside the tags. Each route must independently reach E within the per-route step budget stated above (score is zero if it doesn't).
Format example (m=3):
  <route_1>RIGHT RIGHT RIGHT RIGHT DOWN DOWN DOWN DOWN</route_1>
  <route_2>DOWN DOWN DOWN DOWN RIGHT RIGHT RIGHT RIGHT</route_2>
  <route_3>RIGHT DOWN RIGHT DOWN RIGHT DOWN RIGHT DOWN</route_3>
\end{promptbox}

\subsection{MuSiQue}
\label{app:prompts_musique}

\begin{promptbox}{MuSiQue: single-answer prompt}
Read the paragraphs below and answer the multi-hop question. Identify which paragraphs support your answer.

Paragraphs:
{paragraphs_text}

Question: {question}

First reason about which paragraphs are relevant, then output your supporting paragraph indices and answer.
<support>comma-separated paragraph indices (e.g., 3, 7, 12)</support>
<answer>your answer</answer>
\end{promptbox}

\noindent\texttt{\{paragraphs\_text\}} lists the 20 paragraphs, one per line, as \texttt{[\{idx\}] (Title: \{title\}) \{paragraph\_text\}}.

\paragraph{Multi-answer rewrite.} The last three lines (from ``First reason \ldots'') are replaced by:

\begin{promptbox}{MuSiQue: multi-answer rewrite}
Give {m} different answers in <response_i>...</response_i> tags, each with <support>indices</support> and <answer>answer</answer>.
\end{promptbox}

\subsection{EUREQA}
\label{app:prompts_eureqa}

\begin{promptbox}{EUREQA: single-answer prompt}
You will be given a multi-hop reasoning question over a small narrative. The narrative contains masked entities (Person A, Country B, etc.) connected by relations. The leaf entity is revealed in the final sentence; you must back-chain to identify the rest.

Narrative:
{narrative}

The masked entities you must resolve (in order of introduction in the narrative):
{mask_list}

For each masked entity above, output its underlying canonical name. Use underscored Wikipedia-style names exactly as they appear in the source (e.g. `Robert_Rodriguez`, `32nd_European_Film_Awards`, `From_Dusk_till_Dawn:_The_Series`).

Replace each `...` below with the resolved entity (NOTHING else inside the tags -- just the canonical name):
<entity_A>...</entity_A>   (resolves {mask_a})
<entity_B>...</entity_B>   (resolves {mask_b})
<entity_C>...</entity_C>   (resolves {mask_c})
<entity_D>...</entity_D>   (resolves {mask_d})
<entity_E>...</entity_E>   (resolves {mask_e})

The answer to the question is whatever you put in <entity_A>.
\end{promptbox}

\noindent\texttt{\{mask\_list\}} lists the five masks, one per line, each preceded by ``\texttt{\ \ -\ }''.

\paragraph{Multi-answer rewrite.} The block from ``Replace each \ldots'' to the end is replaced by:

\begin{promptbox}{EUREQA: multi-answer rewrite}
Provide {m} different reasoning chains, each wrapped in numbered tags <response_1>...</response_1> through <response_{m}>...</response_{m}>. Each chain must contain all 5 entity resolutions; replace each `...` with ONLY the resolved canonical name (underscored Wikipedia-style, no prose, no labels):
  <response_i>
    <entity_A>...</entity_A>   (resolves Person A -- this is the answer)
    <entity_B>...</entity_B>   (resolves the second mask)
    <entity_C>...</entity_C>   (resolves the third mask)
    <entity_D>...</entity_D>   (resolves the fourth mask)
    <entity_E>...</entity_E>   (resolves the fifth mask)
  </response_i>
Closing tags are required for every response and every entity. The {m} chains should be genuinely different attempts (e.g. different candidate resolutions, not paraphrases). The answer to the question is whatever you put in <entity_A> within each chain.
\end{promptbox}

\subsection{ToolRL}
\label{app:prompts_toolrl}

ToolRL prompts are used as released with the ToolRL training set: the system message describes the tools and the output format, and the user message contains the dialogue.

\begin{promptbox}{ToolRL: system message}
You are a helpful multi-turn dialogue assistant capable of leveraging tool calls to solve user tasks and provide structured chat responses.

**Available Tools**
In your response, you can use the following tools:
1. Name: <tool_1_name>
Description: <...>
Parameters: {...}
2. Name: <tool_2_name>
...

**Steps for Each Turn**
1. **Think:** Recall relevant context and analyze the current user goal.
2. **Decide on Tool Usage:** If a tool is needed, specify the tool and its parameters.
3. **Respond Appropriately:** If a response is needed, generate one while maintaining consistency across user queries.

**Output Format**
<think> Your thoughts and reasoning </think>
<tool_call>
{"name": "Tool name", "parameters": {"Parameter name": "Parameter content", "... ...": "... ..."}}
{"name": "... ...", "parameters": {"... ...": "... ...", "... ...": "... ..."}}
...
</tool_call>
<response> AI's final response </response>

**Important Notes**
1. You must always include the `<think>` field to outline your reasoning. Provide at least one of `<tool_call>` or `<response>`. Decide whether to use `<tool_call>` (possibly multiple times), `<response>`, or both.
2. You can invoke multiple tool calls simultaneously in the `<tool_call>` fields. Each tool call should be a JSON object with a "name" field and an "parameters" field containing a dictionary of parameters. If no parameters are needed, leave the "parameters" field an empty dictionary.
3. Refer to the previous dialogue records in the history, including the user's queries, previous `<tool_call>`, `<response>`, and any tool feedback noted as `<obs>` (if exists).
\end{promptbox}

\begin{promptbox}{ToolRL: example user message}
**Dialogue Records History**
<user> I need to make sure my library visit is smooth. Could you check if I'm a member, authorize my entry, log the access, and also verify the access control settings at the Main Library for the afternoon? </user>
\end{promptbox}

\paragraph{Multi-answer rewrite.} The system message is unchanged, and the block below is appended to the user message.

\begin{promptbox}{ToolRL: multi-answer rewrite}


---
ADDITIONAL INSTRUCTION (overrides the single-attempt output format above): Provide {m} different attempts at the task. Wrap each attempt in numbered outer tags <response_1>...</response_1> through <response_{m}>...</response_{m}>. Inside each <response_i>, follow the original output format from the system prompt (your <think>, optional <tool_call>, and inner <response> sections). The {m} attempts should be genuinely different -- different tool choices, different argument values, or different reasoning -- not paraphrases. Closing tags are required on every outer attempt.
\end{promptbox}

\subsection{Goal-conditioning suffix}
\label{app:prompts_goalcond}

Goal-conditioned GRPO appends the block below to the single-answer prompt. Each training example has its own weighting $w\sim\mathrm{Dir}(\mathbf{1})$, drawn once and kept for all epochs; at evaluation the block contains $w^\star$ or a fresh draw (App.~\ref{app:method_settings}).

\begin{promptbox}{Goal-conditioning suffix}


Your objective weights (higher = more important to optimize):
- {desc_1}: {w_1:.2f}
- {desc_2}: {w_2:.2f}
...

Maximize your weighted score.
\end{promptbox}

\noindent The descriptions name the reward components. For Maze they are ``Reaching the exit (E)'', ``Collecting Gold (G) tiles'', ``Collecting Diamond (D) tiles'', and ``Avoiding Lava (L) tiles''.

\section{LLM Usage}
\label{app:llm_usage}

We used Anthropic's Claude and OpenAI's GPT-5 for two non-methodological purposes: (i) writing assistance, i.e., drafting, editing, restructuring, and copy-editing prose throughout the manuscript, including this appendix; and (ii) coding assistance for training scripts, evaluation pipelines, plotting code, and reproducibility tooling. LLMs were not used to generate research ideas, theoretical results, experimental designs, or empirical claims. The VPO method, the experimental design, and the analysis are the authors' own.

\end{document}